\documentclass{article}

\usepackage{iclr2023_conference}

\usepackage{soul}
\usepackage[utf8]{inputenc} 
\usepackage[T1]{fontenc}    
\usepackage{hyperref}       
\usepackage{url}            
\usepackage{booktabs}       
\usepackage{amsfonts}       
\usepackage{nicefrac}       
\usepackage{microtype}      
\usepackage{xcolor}         
\usepackage{natbib}
\usepackage{amsmath}
\usepackage{booktabs}
\usepackage{import}
\usepackage{mathtools}
\usepackage{multirow}
\usepackage[labelfont=bf]{caption}
\usepackage{tabularx}
\usepackage{subcaption}
\usepackage{graphicx}
\usepackage[graphicx]{realboxes}
\usepackage[ruled,vlined, linesnumbered]{algorithm2e}
\setcitestyle{square}

\newcommand{\comment}[1]{}

\SetKwComment{Comment}{/* }{ */}
\DeclareMathOperator*{\ReLU}{ReLU}

\DeclareMathOperator*{\N}{\mathbb{Z}}
\def\*#1{\mathbf{#1}}

\DeclareMathOperator*{\R}{\mathbb{R}}
\DeclareMathOperator*{\argmax}{arg\,max}

\def\*#1{\mathbf{#1}}

\title{Pathfinding Neural Cellular Automata}

\author{
\\
\begin{tabularx}{\textwidth}{XX}
  \textbf{Sam Earle} \hfill \textbf{Ozlem Yildiz} \\
  NYU Tandon \hfill NYU Tandon \\
  \texttt{sam.earle@nyu.edu} \hfill \texttt{ozlem.yildiz@nyu.edu} \\
  \\
  \textbf{Julian Togelius} \hfill \textbf{Chinmay  Hegde}\\
  NYU Tandon \hfill NYU Tandon \\
  \texttt{julian.togelius@nyu.edu} \hfill  \texttt{chinmay.h@nyu.edu} \\  
\end{tabularx}}

\begin{document}

\maketitle

\begin{abstract}
Pathfinding makes up an important sub-component of a broad range of complex tasks in AI, such as robot path planning, transport routing, and game playing. 
While classical algorithms can efficiently compute shortest paths, neural networks could be better suited to adapting these sub-routines to more complex and intractable tasks.
As a step toward developing such networks, we hand-code and learn models for Breadth-First Search (BFS), i.e. shortest path finding, using the unified architectural framework of Neural Cellular Automata, which are iterative neural networks with equal-size inputs and outputs. 
Similarly, we present a neural implementation of Depth-First Search (DFS), and outline how it can be combined with neural BFS to produce an NCA for computing diameter of a graph. We experiment with architectural modifications inspired by these hand-coded NCAs, training networks from scratch to solve the diameter problem on grid mazes while exhibiting strong generalization ability. 
Finally, we introduce a scheme in which data points are mutated adversarially during training.
We find that adversarially evolving mazes leads to increased generalization on out-of-distribution examples, while at the same time generating data-sets with significantly more complex solutions for reasoning tasks.\footnote{Code is available at 
\hyperlink{https://anonymous.4open.science/r/pathfinding-nca-FEAD}{https://anonymous.4open.science/r/pathfinding-nca-FEAD}
}

\end{abstract}

\section{Introduction}

Pathfinding is a crucial sub-routine in many important applications. On a 2D grid, the shortest path problem is useful for robot path planning (\cite{robotpath}) or in transportation routing (\cite{trasportpath}). Long paths (and quantities such as the diameter) are relevant for estimating photovoltaic properties of procedurally-generated microstructures for solar panels (\cite{stenzel2016predicting}, \cite{lee2021fast}), or the complexity of grid-based video game levels (\cite{earle2021illuminating}).

Classical algorithms to solve pathfinding and related problems include the Bellman-Ford algorithm~(\cite{bellman1958routing} \cite{ford1956network}) for finding the shortest path from a single source node to other nodes, Breadth-First Search (BFS) (\cite{moore1959shortest}, \cite{merrill2012scalable}), which models the connected nodes using a Dijkstra map (\cite{dijkstra1959note}), and  Depth-First Search (DFS)~(\cite{tarjan1972depth}), which explores the connections from each node sequentially.

Neural networks are increasingly being used for solving complex problems in the aforementioned applications involving pathfinding subroutines. Therefore, modeling classical pathfinding algorithms in ``neurally plausible''  ways could be advantageous for holistically solving these more complex problems. This approach has been explored before: for the shortest path problem, \cite{kulvicious2021withouttraining} construct hand-crafted neural networks to implement an efficient, distributed version of BFS.

On the other hand, we also know that the performance and generalization of neural networks depend heavily on their structure. \cite{xu2019can} posit the theory of \emph{algorithmic alignment}, which is a measure of a network architecture's appropriateness for a reasoning task. 
If the network structure aligns with an algorithm for solving the target reasoning task, the network's sample complexity is lower. 
For example, the structure of the Bellman-Ford algorithm for shortest path finding aligns with Graph Neural Networks (GNNs) more than Multi-Layer Perceptrons (MLPs), and indeed GNNs are shown to generalize well on this task. More specifically, Definition 3.4 of \cite{xu2019can} asserts that networks are aligned when sub-modules of the network have a natural mapping onto sub-functions of the reasoning algorithm (e.g. when it is sufficient for each network submodule to learn the operation inside a for-loop in the target algorithm, instead of the for-loop itself).



\begin{figure}
\begin{subfigure}{0.2\textwidth}
\includegraphics[width=.96\textwidth,trim={0 0 35 0},clip]{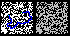}
\caption{}
\label{fig:maze1}
\end{subfigure}
\begin{subfigure}{0.8\textwidth}
\includegraphics[width=.24\textwidth,trim={35 0 0 0},clip]{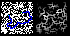}    
\includegraphics[width=.24\textwidth,trim={35 0 0 0},clip]{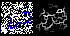}    
\includegraphics[width=.24\textwidth,trim={35 0 0 0},clip]{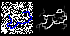}    
\includegraphics[width=.24\textwidth,trim={35 0 0 0},clip]{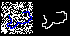}    
\caption{}
\label{fig:shortpath}
\end{subfigure}
    \caption{\textbf{Learned pathfinding behavior.} a) An example maze with the shortest path, where blue, green, dark blue, and black respectively represent source, target, path, and wall tiles. b) A learned model computes the shortest path on an out-of-distribution example. It first activates all traversable tiles, then strengthens activations between source and target, while gradually pruning away the rest.}
    \label{fig:pf_maze}
\vspace*{-0.5 cm}
\end{figure}

\textbf{Problem formulation.} We focus on pathfinding in grid-based mazes with obstacles and empty tiles. While navigating the maze, one can move up, down, right, and left onto empty tiles. We consider two pathfinding problems: finding (i) the shortest path between fixed source and target (Fig.~\ref{fig:maze1}), and (ii) the diameter, which is the longest shortest path between any pair of nodes.



An optimal method for the shortest path problem involves BFS, while an optimal method for the diameter problem includes BFS and DFS (\cite{holzer2012optimal}). We follow the above line of work by implementing target algorithms as neural networks, and using them to propose architectural modifications to learning networks. We conceptualize \textit{architectural alignment} as adjustments to network sub-modules that facilitate their ability to learn sub-functions of the target algorithm, when the mapping between sub-modules/functions is fixed.

Specifically, we showcase the architectural alignment of Neural Cellular Automata (NCAs) (\cite{mordvintsev2020growing})---consisting of a single convolutional layer that is repeatedly applied to an input to iteratively produce an output of the same size---for pathfinding problems on grid-based mazes. NCAs are a natural choice for grid-based domains; like GNNs, they involve strictly local computation and are thus well-suited to similar problems. By the theoretical framework of \cite{xu2019can} (Definition 3.4), the lesser sample complexity of GNNs for pathfinding should apply equally to NCAs.



A summary of our contributions is as follows:
\begin{itemize}
    \item We develop hand-coded NCAs for the shortest path finding problem (Sec.~\ref{s:computepath}), which implement Dijkstra map generation (Sec.~\ref{s:dijks}) and path extraction (Sec.~\ref{s:extract}). The latter is also necessary to extract the shortest path from the Dijkstra map generated by \cite{kulvicious2021withouttraining}. We thus demonstrate that NCAs can complete all the necessary sub-tasks for the shortest path problem.
    
    \item We provide an NCA implementation of DFS (Sec.~\ref{s:dfs}), an essential component in the optimal, parallelized diameter-computing algorithm introduced by \cite{holzer2012optimal}, and outline how BFS- and DFS-NCAs can be combined to solve the diameter problem (Sec.~\ref{s:computediameter}).

    \item We suggest that NCA architectures can be further manipulated to align with the structure of hand-coded solutions for the problem in order to improve their performance, and we support our handcoding-inspired architectural modifications (Sec.~\ref{s:alignment}) with experiments in Sec.~\ref{s:experiments}.
\end{itemize}

Beyond motivating architectural alignment, these differentiable hand-coded models could act as fixed submodules or initialization schemes for larger hybrid architectures, potentially leading to more accurate and reliable learned solutions to more complex problems involving pathfinding.

It is worth noting that NCAs have been generalized to graphs by \cite{grattarola2021learning} as Graph-NCAs, which can be seen to have at least the same representational capacity as convolutional NCAs on grids, while at the same time being applicable to arbitrary graphs.
The GNNs trained on grid-maze domains in Tables~\ref{tab:pf_models}, \ref{tab:app_gnn_grid_rep}, and \ref{tab:app_gnn_archs}, are effectively Graph-NCAs.
Using the message passing layer of, e.g., Graph Attention Networks (GATs, \cite{velivckovic2018graph}, studied in Table~\ref{tab:app_gnn_archs}), Graph-NCAs can compute anisotropic filters, which are crucial for solving the diameter problem.
Future work could investigate the performance of pathfinding-related problems on general graphs using Graph-NCAs.




\vspace*{-0.2cm}
\section{Hand-coded networks}
\label{s:handcode}
We consider two problems: finding theshortest path and diameter. In Sec~\ref{s:dijks}, we use NCAs to implement the generation of the Dijkstra activation map 
since it is the core component in finding the shortest path. 
In Sec~\ref{s:extract} we implement path extraction using an NCA so that the shortest path can be extracted from the Dijkstra map using NCAs.

The optimal algorithm to find the diameter (\cite{holzer2012optimal}) includes DFS, Dijkstra map generation, and path extraction. For the shortest path problem, we provide neural implementations of Dijkstra map generation and path-extraction, and in Sec~\ref{s:dfs} we present a hand-coded DFS NCA. We outline a neural implementation of the diameter algorithm using our hand-coded neural networks in Alg. \ref{alg:1}, which we describe in Sec.~\ref{s:computediameter}.


We design our experiments in Sec~\ref{s:experiments} according to the modifications suggested by our hand-coded implementations, discussed in Sec~\ref{s:alignment}. 
A detailed description of the hand-coded parameters and forward pass is left for Appendix~\ref{s:handcodeappendix}.

    



\vspace*{-0.2cm}
\subsection{Dijkstra map generation}
\label{s:dijks}

We present our hand-coded NCAs to replicate the Dijkstra activation sub-process of BFS. Our approach is similar to \cite{kulvicious2021withouttraining}, though we do not use max-pooling (relying only on $\mathrm{ReLU}$ activations), and we implement bi-directional BFS. We denote the model by $\mathrm{Dijkstra} (u)$ for the source node $u$, and present the full implementation details in \ref{s:handcodeappendix_dijkstra}. 


Our model maintains the following key channels: $\mathrm {flood}_s$ ($\mathrm{flood}_t$) to denote if there is any possible path from the source (target) to the current node, and $\mathrm{age}$ to count the number of iterations for which any $\mathrm{flood}$ has been active at that tile. 
The idea is to flood any empty (i.e. non-$\mathrm{wall}$) tile once an adjacent tile is flooded.
Flood activations are binary-valued to represent if the flood has reached this tile from source/target;
therefore, the activation is $\mathrm{step}$ function, which can be implemented using $\mathrm{ReLU}$s. 
The $\mathrm{age}$ activation is integer-valued, starting at 0, and is incremented whenever the corresponding tile is flooded; therefore, a $\mathrm{ReLU}$ is the activation of this channel. Our model is bi-directional because the $\mathrm{flood}$ activations propagate from both source and target. We use the $\mathrm{age}$ channel to reconstruct the optimal path in our shortest path  extraction NCA after ${\rm flood}_s$ and $\mathrm{flood}_t$ are connected.


\vspace*{-0.2cm}
\subsection{Shortest path extraction}
\label{s:extract}

The Dijkstra activation map includes all tiles reached by either source or target floods, but one needs to identify the shortest path itself. The path can be reconstructed starting from the point(s) at which the two floods meet.  
(Note that there may be more than one possible shortest path and our model will simultaneously extract all possible shortest paths to the $\mathrm{path}$ channel, though ties could be broken using a similar directional queuing mechanism as will be described in our implementation of DFS.)

The flood variables ${\rm flood}_s$ and ${\rm flood}_t$ will meet in the middle of the shortest path, so when ${\rm flood}_s$ $\times$ ${\rm flood}_t$ becomes a positive number, the $\mathrm{path}$ channel will be active and path extraction will begin.
There are four different channels, ${\rm path}_{i, j}$, with $(i,j) \in \{  (1,2),(1,0), (0,1), (2,1) \}$, to detect the presence of $\mathrm{path}$ activation at adjacent tiles to the right, left, top, and bottom, respectively.  

To determine if a tile $u$ should be included in the path being extracted, we must check that one of its neighbors, $v$, is already included in the path, and that $u$ was flooded directly prior to $v$ during the Dijkstra activation.
Therefore, assuming $v$ is the $(i,j)^\text{th}$ neighbor of $u$, then ${\rm path}_{i,j}$ is active when the age of $u$, ${\rm age}_u$, is equal to ${\rm age}_v +1$ and ${\rm path}_v = 1$. 
We use a $\mathrm{sawtooth}$ to determine when these conditions are met at ${\rm path}_{i,j}$. 
Finally, we determine $\mathrm{path}$ channel of a given tile by summing over its ${\rm path}_{i,j}$ channels and passing through the $\mathrm{step}$ function. We denote this model by $\mathrm{PathExtract} (\cdot)$ with the input of Dijkstra map generated by $\mathrm{Dijkstra}(u)$ for the source node $u$. 


\subsection{Depth-First Search}
\label{s:dfs}

While BFS operates in parallel, DFS must run sequentially. Therefore, it will use a stack, with last-in, first-out order, to store nodes that are on the unexplored ``frontier,'' and return a node from the stack when the currently explored route reaches a dead end.


In order to flood the search tree one tile at a time, we prioritize neighbors based on their relative position. In our implementation, the priority decreases in the order of moving down, right, up, and left. We represent the sequential flood in the $\mathrm{route}$ channel, which is binary-valued. Similar to the path extraction model, we keep the directional routes in different channels as denoted in ${ \rm route}_{i,j}$ with $(i,j) \in \{(1,2), (2,1), (3,2), (2,3)\}$, representing down, right, up  and left moves and the main sequential route in $\mathrm{route}$.

In this model, the first activation of the $\mathrm{route}$ is due to the source channel, and then any center tile's $\mathrm{route}$ is activated when it's the priority node of the neighbor with an active $\mathrm{route}$. Therefore, the kernel size will change to $5\times 5$, so the nodes can consider their neighbors' neighbors in order to determine whether they have priority in receiving an adjacent activation. The activation for ${\rm route}_{(i,j)}$ is given by a ReLU, and we calculate $\mathrm{route}$ by summing over ${\rm route}_{(i,j)}$ channels with the $\mathrm{step}$ activation, which is similar to directional paths,  ${\rm path}(i,j)$.

We've shown that the $\mathrm{route}$ builds the DFS search tree. Here, we need to make sure that we stack our possible branches to continue from a different route when the current sequential route is stuck. Therefore, we define a binary-valued $\mathrm{pebble}$ channel to follow the current $\mathrm{route}$, and we can use the existence of $\mathrm{pebble}$ to determine whether we need to pop the last-in tile from the stack. In order to represent the stack model, we declare three different channels: $\mathrm{stack}$, which is binary-valued to denote if the tile is in the stack, $\mathrm{stack \ direction \ chan}$, which is $\in \{0.2, 0.4, 0.6,0.8\}$ to keep the priority for the tiles that are added to the stack, and $\mathrm{stack \ rank \ chan}$, which is integer-valued to count the iteration time to keep them in the general order. 

We note that the $\mathrm{pebble}$ channel is calculated by $\mathrm{route^{t}} - \mathrm{route^{t-1}}$ where $t$ is the current timestep. 
We define a $since$ channel to represent the number of timesteps that have passed since $\mathrm{pebble}$ was active on the adjacent tile (which is added to the stack).



The activation of the $\mathrm{stack}$ happens when max-pooling the $\mathrm{pebble}$ channel over the entire maze returns $0$, which means that $\mathrm{route}$ is stuck. This activates the min-pooling over the entire maze of $\mathrm{stack \ rank} + \mathrm{stack \ direction}$ (excluding tiles where this sums to $0$). 
Min-pooling is used to determine the tile that was most recently added to the stack. This tile's $\mathrm{stack}$ activation is changed to 0, clearing the way for $\mathrm{route}$ activation.
(If the same tile is added to the stack twice, we use a sawtooth to identify if any tile has $\mathrm{stack \ chan}= 2$, in which case we use skip connections to \textit{overwrite} previous $\mathrm{stack \ rank}$ and $\mathrm{direction}$ activations.)

We emphasize that the DFS scheme outlined here can be implemented using convolutional weights, skip connections, max-pooling, and ReLU activations, and is thus differentiable. We denote the operation of this hand-coded neural network by $\mathrm{DFS}(u)$ for a starting node $u$.

\section{Pathfinding by hand-coded neural networks}

\subsection{Computing the shortest path}
\label{s:computepath}

Our goal is to compute the shortest path in the given maze when there is a defined source $u$. Thus, we first generate a Dijkstra map by the function call of $\mathrm{Dijkstra}(u)$, then, we extract the path from this Dijkstra map by calling $\mathrm{PathExtract}$ over this Dijkstra map. 
We can thus denote the shortest path finding routine conducted by our handcoded neural networks as $\mathrm{PathExtract} (\mathrm{Dijkstra} (\mathrm{u}))$.
In practice, the networks' convolutional weights are concatenated, then the non-linearities and other computations specific to each routine are applied to the result in sequence.

\subsection{Computing the diameter}
\label{s:computediameter}

In order to calculate the diameter, one could naively calculate the shortest path between all pairs of nodes and extract the largest path among them in $O(n^3)$ time. An asymptotically near-optimal method for calculating the diameter has been proposed by \cite{holzer2012optimal}, and includes BFS and DFS as subroutines and runs in $O(n)$. 
This algorithm relies on a parallelisable message-passing scheme.

In Alg. \ref{alg:1}, we propose a neural implementation of the diameter algorithm. We assume that all nodes are connected in the graph. (Otherwise, the diameter routine would need to be called on an arbitrary node from each connected component.)

The algorithm starts with the $\mathrm{DFS}$ call on an initial randomly chosen node. While $\mathrm{DFS}$ is running, we make a call to the Dijkstra map generation routine, $\mathrm{Dijkstra}(u)$, whenever $\mathrm{pebble}$ is active at $u$. 

To prevent the collision of the flooding frontiers from parallel Dijkstra routines, we must wait a certain number of timesteps before each new call. 
The $\mathrm{pebble}$ activation ``moves'' between timesteps in two ways: either directly between two adjacent tiles, or by jumping across tiles to the last on the $\mathrm{stack}$ because no adjacent tiles were visitable. In the first case, the waiting time is one, so that the new call to $\mathrm{Dijkstra}$ starts after the first step of the previous call. In the latter case, it needs to backtrack to the next node from the $\mathrm{stack}$, so the waiting time should be equal to backtracking time $+1$. 
This ensures that each new flood is ``inside'' the previous one so there is  no collision

Each cell $u$ in the grid stores its $\mathrm{age}$ as $\mathrm{path\_max}$ when $\mathrm{Dijkstra}(u)$ returns, since in our $\mathrm{Dijkstra}$ implementation, the source has the largest age.
When the $\mathrm{DFS}$ routine is completed, each $\mathrm{path\_max}$ corresponds to the longest shortest path in which the given tile is an endpoint.
We calculate the diameter by taking the largest $\mathrm{path\_max}$, then extract the corresponding shortest path.


 \vspace*{-0.2cm}
\begin{algorithm}
\DontPrintSemicolon
\SetNoFillComment

\caption{Diameter }\label{alg:1}
$i=0$, $\mathrm{prior} = \mathrm{null}$, $\mathrm{waiting\_time} = 0$, Choose a node $v$, start $\mathrm{DFS}(v)$\;
\While{$\mathrm{DFS}$ runs}{
\If {$u^{\mathrm{pebble}} == 1$}{
    \lIf {$\mathrm{prior}\ != null$}{
        $ \mathrm{waiting\_time} = \left| \mathrm{prior}^{\mathrm{since}} - u^{\mathrm{since}}\right|$\; 
    }
    \vspace*{-0.4cm}
     $\mathrm{prior} = u$,  \textit{Wait} $\mathrm{waiting\_time}$, $\mathrm{Dijkstra}(u)$\; 

    \lIf{ $\mathrm{Dijkstra}(v)$ ends}{
        $v^{\mathrm{path\_max}} =  v^{\mathrm{age}},$\;
        \vspace*{-0.4cm}
    }
       
}
}
$\mathrm{PathExtract} (\mathrm{Dijkstra} (\argmax (\mathrm{path\_max})))$\;
\end{algorithm}
 \vspace*{-0.6cm}
\section{Architectural modifications}
\label{s:alignment}

\begin{table}
\resizebox{\linewidth}{!}{

\begin{tabular}{|c|r|r|r|r|r|r|r|}
\hline
    &    &     &       model &                  train & \multicolumn{3}{c|}{test} \\
\cline{4-8}
\cline{4-8}
    &    &     &         --- &                  16x16 & \multicolumn{2}{c|}{16x16} &                  32x32 \\
\cline{4-8}
\cline{4-8}
    &    &     &   n. params &             accuracies &             accuracies &          pct. complete &             accuracies \\
\textbf{model} & \textbf{n. layers} & \textbf{n. hid chan} &             &                        &                        &                        &                        \\
\hline
\multirow{2}{*}{\textbf{GCN}} & \multirow{2}{*}{\textbf{32}} & \textbf{96 } &       9,600 &          37.89 ± 34.59 &          37.95 ± 34.65 &          18.24 ± 16.65 &          25.01 ± 23.75 \\

    &    & \textbf{256} &      66,560 &          69.61 ± 38.92 &          69.36 ± 38.78 &          41.09 ± 22.97 &          43.92 ± 27.12 \\
\cline{2-8}
    
\cline{1-8}
\cline{2-8}
\multirow{2}{*}{\textbf{MLP}} &  \multirow{2}{*}{\textbf{64}} & \textbf{96 } &  16,257,024 &           70.69 ± 3.29 &           13.91 ± 1.04 &            2.74 ± 0.20 &            0.00 ± 0.00 \\
    &    & \textbf{256} &  42,799,104 &           52.92 ± 6.63 &           12.82 ± 1.78 &            1.94 ± 0.17 &            0.00 ± 0.00 \\
\cline{1-8}
\cline{2-8}
\multirow{2}{*}{\textbf{NCA}} 
    & \multirow{2}{*}{\textbf{32}} & \textbf{96 } &      86,400 &           \textbf{99.67}± 0.28 &           \textbf{96.79} ± 0.70 &           \textbf{96.26} ± 0.47 &          \textbf{84.75} ± 6.69 \\

    &    & \textbf{256} &     599,040 &          79.68 ± 44.54 &          78.33 ± 43.79 &          78.27 ± 43.76 &          74.24 ± 41.55 \\
\hline
\end{tabular}
}
\caption{\textbf{Shortest path problem -- model architecture} : For each architecture, we choose the minimal number of layers capable of achieving high generalization. NCAs with weight sharing generalize best, while GCNs generalize better than MLPs. }
\label{tab:pf_models}
\vspace*{-0.5cm}
\end{table}

\begin{table}
\resizebox{\linewidth}{!}{
\begin{tabular}{|c|r|r|r|r|r|r|r|}
\hline
    &       &     &      model &                  train & \multicolumn{3}{c|}{test} \\
\cline{4-8}
\cline{4-8}
    &       &     &        --- &                  16x16 & \multicolumn{2}{c|}{16x16} &                  32x32 \\
\cline{4-8}
\cline{4-8}
    &       &     &  n. params &             accuracies &             accuracies &          pct. complete &             accuracies \\
\textbf{model} & \textbf{shared weights} & \textbf{n. hid chan}  &            &                        &                        &                        &                        \\
\hline
\multirow{8}{*}{\textbf{NCA}} & \multirow{4}{*}{\textbf{False}} & \textbf{32 } &    663,552 &           99.73 ± 0.31 &           94.96 ± 0.88 &           94.37 ± 0.56 &          58.95 ± 40.94 \\
    &       & \textbf{48 } &  1,437,696 &          59.90 ± 54.68 &          57.49 ± 52.49 &          57.26 ± 52.27 &          49.23 ± 44.96 \\
    &       & \textbf{96 } &  5,529,600 &  \textbf{99.74} ± 0.28 &           96.42 ± 1.09 &           96.90 ± 0.37 &           81.79 ± 4.36 \\
    &       & \textbf{128} &  9,732,096 &          79.91 ± 44.67 &          77.65 ± 43.41 &          77.55 ± 43.35 &          67.83 ± 38.01 \\
\cline{2-8}
    & \multirow{4}{*}{\textbf{True }} & \textbf{32 } &     10,368 &          58.09 ± 53.03 &          56.80 ± 51.86 &          56.54 ± 51.62 &          49.44 ± 45.16 \\
    &       & \textbf{48 } &     22,464 &           98.68 ± 0.44 &  \textbf{96.61} ± 0.33 &  \textbf{96.92} ± 0.09 &  \textbf{88.83} ± 1.78 \\
    &       & \textbf{96 } &     86,400 &          78.68 ± 43.99 &          77.64 ± 43.41 &          78.05 ± 43.64 &          72.21 ± 40.42 \\
    &       & \textbf{128} &    152,064 &          39.57 ± 54.18 &          39.25 ± 53.75 &          39.38 ± 53.92 &          36.94 ± 50.58 \\
\hline
\end{tabular}
}
\caption{\textbf{Shortest path problem -- weight sharing}: Sharing weights between layers can improve generalization while reducing the number of learnable parameters.}
\label{tab:weight_share}
\vspace*{-0.3cm}
\end{table}

\begin{table}
\resizebox{\linewidth}{!}{
\begin{tabular}{|c|r|r|r|r|r|r|r|}
\hline
    &       &     &     model &                  train & \multicolumn{3}{c|}{test} \\
\cline{4-8}
\cline{4-8}
    &       &     &       --- &                  16x16 & \multicolumn{2}{c|}{16x16} &                  32x32 \\
\cline{4-8}
\cline{4-8}
    &       &     & n. params &             accuracy &             accuracy &          pct. complete &             accuracy \\
\textbf{model} & \textbf{cut corners} & \textbf{n. hid  chan}  &           &                        &                        &                        &                        \\
\hline

\multirow{6}{*}{\textbf{NCA}} & \multirow{3}{*}{\textbf{False}} & \textbf{48 } &    22,464 &           98.68 ± 0.44 &           96.61 ± 0.33 &           96.92 ± 0.09 &           88.83 ± 1.78 \\
    &       & \textbf{96 } &    86,400 &          78.68 ± 43.99 &          77.64 ± 43.41 &          78.05 ± 43.64 &          72.21 ± 40.42 \\
    &       & \textbf{128} &   152,064 &          39.57 ± 54.18 &          39.25 ± 53.75 &          39.38 ± 53.92 &          36.94 ± 50.58 \\
\cline{2-8}
    & \multirow{3}{*}{\textbf{True }} & \textbf{48 } &    12,480 &           97.43 ± 0.88 &           95.81 ± 0.91 &           95.41 ± 1.17 &          81.47 ± 10.00 \\
    &       & \textbf{96 } &    48,000 &  \textbf{98.97} ± 0.32 &  \textbf{97.61} ± 0.31 &  \textbf{97.83} ± 0.50 &  \textbf{90.25} ± 3.88 \\
    &       & \textbf{128} &    84,480 &          79.12 ± 44.23 &          78.34 ± 43.80 &          78.57 ± 43.92 &          75.49 ± 42.22 \\
\hline
\end{tabular}
}
\caption{\textbf{Shortest path problem -- cutting corners}: Ignoring diagonal relationships between grid cells allows for comparable performance with fewer parameters.}
\label{tab:cut_corner}
\vspace*{-0.5cm}
\end{table}

We focus on three sub-tasks via their hand-coded NCA implementations: Dijkstra map generation, path extraction, and DFS. For the shortest path problem, Dijkstra map generation and path extraction are necessary steps. This implementation illustrates the algorithmic alignment of NCAs and also suggests certain architectural modifications to increase their performance. 

In our implementation of shortest path finding (Sec. \ref{s:computepath}), involving Dijkstra map generation and path extraction, we share weights between convolutional layers (each of which can be considered as a single iteration of the algorithm), where the spatial convolutional weights do not have any values in the corners. 
Additionally, we add a skip connection at each layer, feeding in the same one-hot encoding of the original maze, so that the model can reason about the placement of walls (and in particular avoid generating paths that move illegally through them) at each iteration of the algorithm.
We also implement shortest path finding so as to be bi-directional, i.e., flowing out simultaneously from source and target nodes by increasing the number of channels. This halves the number of sequential steps necessary to return the optimal path while adding a small constant number of additional channels. Generally, we hypothesize that more channels could be leveraged---whether by human design or learned models---to make the algorithm return in a fewer number of iterations, offloading sequential operations (distinct convolutional layers) to parallelized ones (additional activations and weights to be processed in parallel).

In Sec~\ref{s:experiments}, we take inspiration from these observations and implementation tricks, investigating the effect of analogously constraining or augmenting a learning model. In particular, we consider an increased number of channels, weight sharing, 
and alternative convolutional kernel shape, and in several cases observe increased performance on the shortest path problem.


The diameter algorithm includes all three subroutines (Sec. \ref{s:computediameter}). During DFS, we increase the kernel size from $3 \times 3$ to $5\times 5$ to keep track of the edges' priorities. Also, DFS relies on max-pooling as a non-local subroutine, which suggests that it should be included in learned architectures to promote generalization (\cite{xu2020neural}). Therefore, in Sec~\ref{s:experiments}, we investigate the effects of increasing kernel size and adding max-pooling layers to the NCA network on the diameter problem.


\comment{
\begin{table}
\resizebox{\linewidth}{!}{
\begin{tabular}{|c|r|r|r|r|r|r|r|}
\hline
    &      &     &      model & \multicolumn{2}{c|}{train} & \multicolumn{2}{c|}{test} \\
\cline{4-8}
    &      &     &  n. params &           pct. complete &       completion time &          pct. complete &       completion time \\
\textbf{model} & \textbf{shared weights} & \textbf{n. hid chan} &            &                         &                       &                        &                       \\
\hline
\multirow{12}{*}{\textbf{NCA}} & \multirow{6}{*}{\textbf{False}} & \textbf{8  } &     31,232 &            89.99 ± 4.00 &           3.64 ± 5.65 &           72.71 ± 5.28 &           4.02 ± 5.92 \\
    &      & \textbf{16 } &    103,424 &            99.12 ± 0.95 &           1.50 ± 3.98 &           80.32 ± 4.85 &           1.04 ± 3.23 \\
    &      & \textbf{32 } &    370,688 &            80.62 ± 4.60 &           0.90 ± 2.89 &           70.26 ± 5.23 &           0.91 ± 2.88 \\
    &      & \textbf{48 } &    801,792 &  \textbf{100.00} ± 0.00 &           2.18 ± 3.66 &           87.21 ± 2.93 &           2.04 ± 3.65 \\
    &      & \textbf{96 } &  3,078,144 &            99.80 ± 0.52 &  \textbf{0.61} ± 2.27 &           89.40 ± 4.38 &  \textbf{0.65} ± 2.40 \\
    &      & \textbf{128} &  5,414,912 &            99.27 ± 1.17 &           2.18 ± 4.37 &           89.16 ± 3.66 &           2.16 ± 4.42 \\
\cline{2-8}
    & \multirow{6}{*}{\textbf{True}} & \textbf{8  } &        488 &            21.34 ± 5.07 &          10.31 ± 4.37 &           21.73 ± 4.27 &          10.51 ± 4.32 \\
    &      & \textbf{16 } &      1,616 &            73.49 ± 4.79 &           7.88 ± 3.90 &           69.34 ± 5.41 &           7.83 ± 3.91 \\
    &      & \textbf{32 } &      5,792 &            89.40 ± 4.05 &           6.59 ± 3.41 &           85.69 ± 4.42 &           6.59 ± 3.46 \\
    &      & \textbf{48 } &     12,528 &            98.00 ± 2.45 &           6.30 ± 3.01 &           88.57 ± 4.22 &           6.51 ± 3.22 \\
    &      & \textbf{96 } &     48,096 &            98.19 ± 1.84 &           5.88 ± 3.29 &           92.14 ± 3.11 &           6.01 ± 3.38 \\
    &      & \textbf{128} &     84,608 &            99.76 ± 0.69 &           5.89 ± 3.19 &  \textbf{95.21} ± 2.97 &           5.98 ± 3.24 \\
\hline
\end{tabular}
}
\caption{\textbf{Shortest path problem: weight sharing.} Weight sharing between convolutional layers leads to increased performance on test mazes while learning distinct layers allows for lower time complexity during successful pathfinding episodes.}
\label{tab:pf_weight_sharing}
\end{table}
}

\comment{
\begin{table}
\resizebox{\linewidth}{!}{
\begin{tabular}{|c|r|r|r|r|r|r|}
\hline
    &     &     model & \multicolumn{2}{c|}{train} & \multicolumn{2}{c|}{test} \\
\cline{3-7}
    &     & n. params &          pct. complete &       completion time &          pct. complete &       completion time \\
\textbf{model} & \textbf{n. hid chan} &           &                        &                       &                        &                       \\
\hline
\multirow{10}{*}{\textbf{NCA}} & \textbf{6  } &       306 &           10.45 ± 3.50 &  \textbf{3.51} ± 1.61 &            8.94 ± 3.56 &  \textbf{3.06} ± 1.31 \\
    & \textbf{8  } &       488 &           29.59 ± 6.38 &           9.06 ± 3.64 &           31.15 ± 4.81 &           9.20 ± 3.70 \\
    & \textbf{10 } &       710 &           45.65 ± 6.47 &           5.06 ± 4.24 &           43.65 ± 5.40 &           4.89 ± 4.06 \\
    & \textbf{16 } &     1,616 &           78.66 ± 5.68 &           7.13 ± 3.57 &           77.73 ± 5.00 &           6.99 ± 3.58 \\
    & \textbf{24 } &     3,384 &           53.17 ± 5.79 &           6.79 ± 2.91 &           49.51 ± 6.48 &           6.48 ± 2.83 \\
    & \textbf{32 } &     5,792 &           90.19 ± 3.30 &           6.59 ± 3.66 &           84.91 ± 3.74 &           6.63 ± 3.68 \\
    & \textbf{48 } &    12,528 &           92.72 ± 3.98 &           5.88 ± 3.78 &           86.72 ± 3.64 &           6.01 ± 3.79 \\
    & \textbf{96 } &    48,096 &           95.85 ± 2.78 &           6.06 ± 3.21 &           86.87 ± 5.49 &           6.22 ± 3.34 \\
    & \textbf{128} &    84,608 &  \textbf{99.17} ± 1.10 &           6.16 ± 2.95 &  \textbf{92.43} ± 4.06 &           6.28 ± 3.02 \\
    & \textbf{256} &   333,056 &           98.29 ± 1.67 &           6.33 ± 2.91 &           90.77 ± 3.95 &           6.48 ± 3.07 \\
\hline
\end{tabular}
}
\caption{\textbf{Shortest path problem: number of hidden channels.} Increasing the width of the network (i.e. number of hidden channels in each convolutional layer) increases the correctness of returned paths (but not the speed at which they are found).}
\label{tab:pf_hid_chan}
\end{table}
}
\comment{
\begin{table}
\resizebox{\linewidth}{!}{
\begin{tabular}{|c|r|r|r|r|r|r|r|}
\hline
    &      &     &     model & \multicolumn{2}{c|}{train} & \multicolumn{2}{c|}{test} \\
\cline{4-8}
    &      &     & n. params &          pct. complete &       completion time &          pct. complete &       completion time \\
\textbf{model} & \textbf{cut corners} & \textbf{n. hid chan} &           &                        &                       &                        &                       \\
\hline
\multirow{15}{*}{\textbf{NCA}} & \multirow{7}{*}{\textbf{False}} & \textbf{8  } &       872 &           59.52 ± 6.17 &           8.52 ± 3.34 &           58.94 ± 5.56 &           8.46 ± 3.43 \\
    &      & \textbf{16 } &     2,896 &           81.05 ± 4.92 &           6.99 ± 3.59 &           79.15 ± 4.27 &           6.90 ± 3.53 \\
    &      & \textbf{24 } &     6,072 &           88.62 ± 2.96 &           6.62 ± 3.42 &           84.57 ± 3.88 &           6.59 ± 3.49 \\
    &      & \textbf{32 } &    10,400 &           89.94 ± 4.17 &           5.98 ± 3.58 &           83.35 ± 4.11 &           6.15 ± 3.77 \\
    &      & \textbf{48 } &    22,512 &           92.77 ± 3.19 &           6.04 ± 3.54 &           84.08 ± 5.56 &           6.22 ± 3.62 \\
    &      & \textbf{96 } &    86,496 &  \textbf{99.80} ± 0.52 &           6.08 ± 2.99 &           91.21 ± 3.44 &           6.33 ± 3.14 \\
    &      & \textbf{128} &   152,192 &           99.56 ± 0.70 &           6.11 ± 2.93 &           91.02 ± 2.47 &           6.30 ± 3.07 \\
\cline{2-8}
    & \multirow{8}{*}{\textbf{True}} & \textbf{8  } &       488 &           29.59 ± 4.94 &           9.02 ± 3.59 &           31.15 ± 5.31 &           9.17 ± 3.77 \\
    &      & \textbf{10 } &       710 &           45.65 ± 5.80 &  \textbf{5.06} ± 4.21 &           43.65 ± 5.58 &  \textbf{4.90} ± 4.05 \\
    &      & \textbf{16 } &     1,616 &           78.66 ± 5.04 &           7.12 ± 3.60 &           77.73 ± 4.56 &           7.00 ± 3.52 \\
    &      & \textbf{24 } &     3,384 &           53.17 ± 6.22 &           6.78 ± 2.92 &           49.51 ± 6.78 &           6.49 ± 2.89 \\
    &      & \textbf{32 } &     5,792 &           90.19 ± 3.48 &           6.58 ± 3.64 &           84.91 ± 3.46 &           6.64 ± 3.68 \\
    &      & \textbf{48 } &    12,528 &           92.72 ± 3.51 &           5.88 ± 3.78 &           86.72 ± 5.00 &           5.97 ± 3.81 \\
    &      & \textbf{96 } &    48,096 &           95.85 ± 2.64 &           6.06 ± 3.21 &           86.87 ± 4.86 &           6.22 ± 3.35 \\
    &      & \textbf{128} &    84,608 &           99.17 ± 0.96 &           6.15 ± 2.94 &  \textbf{92.43} ± 3.75 &           6.27 ± 3.00 \\
\hline
\end{tabular}
}
\caption{\textbf{Shortest path problem: cutting corners.} Excluding corner cells from convolutional filters increases the generality of solutions generated by a model when searching for paths in a maze where diagonal movement between tiles is not permitted.}
\label{tab:pf_cut_corners}
\vspace*{-0.5cm}
\end{table}
\vspace*{-0.5cm}
}


\section{Experiments}
\label{s:experiments}


\subsection{Experimental setup}

For NCAs, we follow the network architecture of \cite{mordvintsev2021texture}, who train an NCA to imitate the style of texture images. \footnote{We exclude the RGB-specific pre-processing filters used in this work.}
The NCA consists of one convolutional layer with a $3\times 3$ kernel and padding of $1$, followed by a ReLU activation.

For the GCN architecture, we follow the NCA architecture closely, but use a graph convolutional layer in place of a traditional convolutional one. 
To take full advantage of the graph neural architecture, we represent each maze as a sub-grid including nodes and edges between traversable (non-wall) tiles.

The MLP architecture comprises a series of dense layers followed by ReLU connections. First, the input maze (or intermediary activation) is flattened, then passed via a fully connected encoding layer to a smaller (256 node) activation, then back through another fully connected decoding layer to a large activation, which is finally reshaped back into the size of the input.

We compute the shortest path between source and target nodes using BFS and represent the target path as a binary array with the same width and height as the maze. Loss is computed as the mean squared error between a predicted path array and the target path array. 
The model's output is then clipped to be between $0$ and $1$. 

For the sake of evaluation, accuracy (inverse loss) is normalized against all-zero output, which would achieve $\approx 97\%$ accuracy on the dataset. Accuracy can thus be negative when, e.g., a model predicts a path comprising a majority of non-overlapping tiles relative to the true path. Finally, we record whether, after rounding, the output perfectly matches the target path, i.e. the percentage of target paths perfectly completed, or \textit{pct. complete} in tables.


Each model is comprised of a repeated sequence of identically-structured blocks (comprising a convolutional layer, a graph convolutional layer, or an encoder/decoder comprising two dense layers, in the case of NCAs, GCNs, and MLPs, respectively), with or without weight-sharing between them. 

Maze is represented as a 2D one-hot array with 4 or 2 channels for the shortest path or diameter, respectively since there is no source and target in the diameter problem. We then concatenate a zero-vector with the same width and height as the maze and a given number of hidden channels.
After the input passes through each layer of the model, it is returned as a continuous, multi-channel 2D array with a size equal to the input. We interpret an arbitrarily-chosen hidden channel as the model's predicted path output. After the last layer, we compute the mean-squared error loss of the 2D predicted path output with the ground-truth optimal path.

We use mini-batches of size 64. To feed batches to the GCN, we treat the 64 sub-grids containing the input mazes as disjoint components of a single graph. After each mini-batch, we use Adam (\cite{kingma2014adam}) to update the weights of the model. We train for $50,000$ updates for $5$ trials.

The dataset comprises $10 000$ randomly-generated $16\times 16$ mazes, which are generated by randomly placing empty, wall, source, and target tiles until the target is reachable by the source. The resulting paths are relatively simple, their path length is $\approx 9$ tiles. We additionally test models on $32\times 32$ mazes, which are generated as former and their mean path length $\approx 13$ tiles.




We demonstrate the results to demonstrate the comparison of model architectures, the effects of the architectural modifications, and the importance of the data generation in the following subsections. We refer the reader to Sec.~\ref{s:extended results} in Appendix.

\subsection{Model architecture}

In Table \ref{tab:pf_models}, we demonstrate that NCAs outperform GCNs and MLPs, and generalize better than them. This indicates that NCAs are well-aligned with pathfinding problems over grids. Also, GCNs generalize substantially better than MLPs, which fits with past work that has demonstrated the alignment of Graph Neural Networks with pathfinding tasks \cite{xu2019can}.

The relatively poor performance of GCNs may seem at odds with past work by \cite{xu2019can,tang2020towards}. However, we train on a smaller dataset with more complex mazes for a shorter amount of time compared to earlier works. Also, the goal is to recover the optimal path itself as opposed to merely its length so it's a more complicated problem.




We note that our BFS implementation does not distinguish between a node's neighbors at different positions, and could thus be easily be adapted to use a GCN instead of an NCA. To learn this hand-coding, NCAs would need to learn more structure than GCNs, which come with this spatial symmetry built in. However, it is clear from Fig. \ref{fig:pf_maze} that our learned models are not directly performing our handcoded implementation. 
In particular, they appear to propagate slowly-diminishing activations out from the source and target nodes, progressively strengthening the value of nodes that connect source and target while weakening others.
To produce this behavior, it may be important to know which neighbor provided the activation originally to prioritize neighbors on the receiving end.
(Similarly, our hand-coded DFS-NCA uses spatial distinctions between neighbors to prioritize the distribution of activation among them.)
But the GCN trained here is incapable of making these distinctions given that it applies the same weights to each neighbor and aggregates the results.



\subsection{Shortest path}

Following the modification suggested in Sec.~\ref{s:alignment}, we investigate the  performance after \textit{weight sharing} between layers, and ignoring the corner weights, (\textit{cutting corners}). In Table~\ref{tab:weight_share}, we see that weight-sharing leads to the best performance while drastically decreasing the number of parameters.
This agrees with our knowledge of known pathfinding algorithms, which repeatedly apply the same computations. 
Also, we demonstrate that increasing the number of hidden channels improves performance to a certain extent.
This is reflective of a general trend in deep learning in which overparameterization leads to increased performance.

In Table~\ref{tab:cut_corner}, we examine the effect of modifying the kernel to ignore corners in each $3\times 3$ patch, i.e. \textit{cutting corners}.
Across varying numbers of hidden channels, we observe comparable performance with and without this modification, despite having reduced the number of parameters by $4/10$. 
This supports the intuition from our hand-coded BFS implementation and suggests that diagonal neighbors provide little useful information when determining the next state of a given node when finding optimal paths. 




\subsection{Diameter}


In the diameter problem, we again analyze \textit{cutting corners} and \textit{max-pooling}, as well as the effect of \textit{kernel size}. In Table~\ref{tab:diam_max_pool}, we see that max-pooling has a significant effect on the models' performance and generalization on the diameter problem. Spatial max-pooling allows for the global aggregation of information computed locally at disparate points on the map.
In our hand-coded DFS-NCA, max-pooling is used to \textit{pop} frontier nodes from a stack so that we may traverse them in sequence.
In the neural diameter algorithm, it also corresponds to the $\argmax$ of the shortest paths that have been found in different connected components of the grid.

Table \ref{tab:diam_kernel} suggests that simply increasing kernel size in convolutional layers tends to degrade the performance of NCAs on the diameter problem.
This recalls the performance differences between MLP and GNN/NCA architectures observed in Table \ref{tab:pf_models}, in that MLPs, which observe spatially larger parts of the input maze at once are not robust to mazes outside of the training set. 
In one sense, this would seem to go against the intuition suggested by our hand-coded DFS-NCA, which uses $5\times 5$ kernels.
But very little of these larger patches are actually used in our hand-coded weights, and the potentially extraneous information they provide to a learning model may lead it to make spurious correlations.

Accordingly, in Table \ref{tab:diam_kernel} we also examine the effect of \textit{cutting corners} for cells that go unused in our handcoded DFS-NCA (refer to Table \ref{s:dfs} for the weights). This increases the performance of $5\times 5$ kernels despite resulting in fewer learnable parameters.
Surprisingly, we also note that cutting the corner cells in $3\times 3$ kernels similarly improves performance, suggesting that the diameter task may be feasible (or at least more learnable) when focusing on relationships between directly connected nodes, and leaving non-local computations to, e.g. a small number of max-pooling operations.



\begin{table}
\resizebox{\linewidth}{!}{
\begin{tabular}{|c|r|r|r|r|r|r|r|}
\hline
    &       &     &     model &                  train & \multicolumn{3}{c|}{test} \\
\cline{4-8}
\cline{4-8}
    &       &     &       --- &                  16x16 & \multicolumn{2}{c|}{16x16} &                   32x32 \\
\cline{4-8}
\cline{4-8}
    &       &     & n. params &             accuracies &             accuracies &          pct. complete &              accuracies \\
\textbf{model} & \textbf{max-pool} & \textbf{n. hid chan} &           &                        &                        &                        &                         \\
\hline
\multirow{4}{*}{\textbf{NCA}} & \multirow{2}{*}{\textbf{False}} & \textbf{96 } &    47,040 &          77.69 ± 43.43 &          66.48 ± 37.16 &          57.60 ± 32.21 &        -138.22 ± 204.08 \\
    &       & \textbf{128} &    83,200 &          78.05 ± 43.63 &          66.89 ± 37.40 &          58.38 ± 32.65 &           -9.86 ± 12.18 \\
\cline{2-8}
    & \multirow{2}{*}{\textbf{True }} & \textbf{96 } &    47,040 &           93.61 ± 0.41 &           85.74 ± 0.57 &           64.57 ± 2.07 &  \textbf{56.50} ± 35.32 \\
    &       & \textbf{128} &    83,200 &  \textbf{95.72} ± 0.45 &  \textbf{86.54} ± 1.07 &  \textbf{71.82} ± 1.67 &           56.31 ± 26.80 \\
\hline
\end{tabular}
}
\caption{\textbf{Diameter problem -- max-pooling} (with weight sharing and $3\times 3$ kernels): Adding spatial and channel-wise max-pooling operations at each convolutional layer leads to increased generalization to large ($32\times 32$) out-of-distribution mazes when computing the diameter of a maze, which aligns with tasks involving the global aggregation of locally-computed information.}
\label{tab:diam_max_pool}
\vspace*{-0.4cm}
\end{table}

\begin{table}
\resizebox{\linewidth}{!}{
\begin{tabular}{|c|r|r|r|r|r|r|r|}
\hline
    &   &       &     model &                  train & \multicolumn{3}{c|}{test} \\
\cline{4-8}
\cline{4-8}
    &   &       &       --- &                  16x16 & \multicolumn{2}{c|}{16x16} &                   32x32 \\
\cline{4-8}
\cline{4-8}
    &   &       & n. params &             accuracies &             accuracies &          pct. complete &              accuracies \\
\textbf{model} & \textbf{kernel size} & \textbf{cut corners} &           &                        &                        &                        &                         \\
\hline
\multirow{4}{*}{\textbf{NCA}} & \multirow{2}{*}{\textbf{3}} & \textbf{False} &   149,760 &          77.41 ± 43.27 &          67.20 ± 37.57 &          58.32 ± 32.60 &           27.47 ± 52.59 \\
    &   & \textbf{True } &    83,200 &           95.72 ± 0.45 &  \textbf{86.54} ± 1.07 &  \textbf{71.82} ± 1.67 &  \textbf{56.31} ± 26.80 \\
\cline{2-8}
    & \multirow{2}{*}{\textbf{5}} & \textbf{False} &   416,000 &  \textbf{96.62} ± 0.96 &           76.89 ± 0.93 &           60.15 ± 1.78 &           21.85 ± 14.69 \\
    &   & \textbf{True } &   216,320 &           96.30 ± 0.54 &           81.43 ± 0.45 &           68.72 ± 0.59 &            49.74 ± 2.58 \\
\hline
\end{tabular}
}
\caption{\textbf{Diameter problem -- kernel size/shape} (with max-pooling and 128 channels): Smaller kernel shapes generalize best on the diameter task.}
\label{tab:diam_kernel}
\vspace*{-0.2cm}
\end{table}





\subsection{Adversarial data generation}



\begin{table}
\resizebox{\linewidth}{!}{
\begin{tabular}{|c|r|r|r|r|r|r|r|r|}
\hline
    &       &     &     model -- task & \multicolumn{2}{c|}{train} & \multicolumn{3}{c|}{test} \\
\cline{4-9}
\cline{4-9}
    &       &     &       --- & \multicolumn{2}{c|}{16x16} & \multicolumn{2}{c|}{16x16} &                  32x32 \\
\cline{4-9}
\cline{4-9}
    &       &     & n. params &                sol. length &          pct. complete &             accuracies &          pct. complete &             accuracies \\
\textbf{model} &\shortstack{ \textbf{env} \\ \textbf{generation}} & \shortstack{\textbf{n. hid} \\ \textbf{chan}} &           &                        &                        &                        &                        &                        \\
\hline
\multirow{4}{*}{\shortstack{\textbf{NCA -- Shortest} \\ \textbf{path finding}}} & \multirow{2}{*}{\textbf{False}} & \textbf{96 } &    86,400 &            9.02 ± 0.00 &          78.53 ± 43.92 &          77.09 ± 43.10 &          76.84 ± 42.97 &          65.99 ± 37.70 \\
    &       & \textbf{128} &   152,064 &            9.02 ± 0.00 &  \textbf{99.37} ± 0.24 &           97.44 ± 0.13 &           97.61 ± 0.20 &           90.06 ± 2.93 \\
\cline{2-9}
    & \multirow{2}{*}{\textbf{True }} & \textbf{96 } &    86,400 &           21.39 ± 6.99 &          74.06 ± 41.44 &          79.18 ± 44.26 &          79.21 ± 44.28 &          78.12 ± 43.67 \\
    &       & \textbf{128} &   152,064 &  \textbf{23.75} ± 1.36 &           92.92 ± 3.79 &  \textbf{98.88} ± 0.36 &  \textbf{99.34} ± 0.11 &  \textbf{97.17} ± 2.15 \\
\hline
\cmidrule{0-0}
\morecmidrules

\hline
\multirow{4}{*}{\textbf{NCA -- Diameter}} & \multirow{2}{*}{\textbf{False}} & \textbf{96 } &    84,672 &           24.09 ± 0.00 &           71.58 ± 40.13 &          66.15 ± 36.99 &          55.66 ± 31.15 &           -5.20 ± 41.28 \\
    &       & \textbf{128} &   149,760 &           24.09 ± 0.00 &  \textbf{74.04} ± 41.39 &          67.20 ± 37.57 &          58.32 ± 32.60 &           27.47 ± 52.59 \\
\cline{2-9}
    & \multirow{2}{*}{\textbf{True }} & \textbf{96 } &    84,672 &           26.06 ± 1.11 &           33.94 ± 19.12 &          70.47 ± 39.40 &          52.90 ± 29.61 &  \textbf{60.01} ± 33.55 \\
    &       & \textbf{128} &   149,760 &  \textbf{27.43} ± 0.15 &            49.85 ± 4.07 &  \textbf{90.02} ± 0.44 &  \textbf{75.79} ± 1.57 &           30.02 ± 64.80 \\
\hline

\end{tabular}
}
\caption{\textbf{Online adversarial data evolution} (with $3\times 3$ kernels, max-pooling, and weight-sharing): Adversarial evolution of mazes increases models' generalization ability, as well as the solution length of training mazes.}
\label{tab:diameter_evol_ozlem.tex}
\vspace{-0.5cm}
\end{table}

Recall that we randomly generate a maze dataset in previous experiments. We now apply an evolutionary algorithm to incrementally alter the dataset during training. We evolve reasoning problems to maximize learned models' regret (here, their loss against the ground truth), in the same way, \cite{parker2022evolving} evolve game levels. When the model's loss falls below a certain threshold, we randomly select a batch of data points for mutation, apply noise to the data points (changing the state of some uniformly random set of tiles in the maze), and re-compute the ground-truth solution.

We then evaluate the model's performance on these new mazes (without collecting any gradient), ranking them by their fitness (the loss they induce). We then replace any of the least fit data points from the training set with new \textit{offspring} mazes that are more fit. In Table \ref{tab:diameter_evol_ozlem.tex}, we see that adversarially evolving new data points in this way leads to increased generalization on both tasks. Additionally, this tends to increase the complexity of examples in the training set (by a factor of 2.5 on the pathfinding task).


\subsection{Additional results and comparisons}

NCAs tackle the shortest path problem with high accuracy on both training and test sets (Fig.~\ref{fig:pf_maze}).
GCNs often produce outputs that are reasonable at a high level, though they tend to be blurrier (Fig.~\ref{fig:pf_gcn}). These models are slower to learn and do not appear to have converged in most of our experiments. 
MLPs, when confronted with the shortest path problem, will often reproduce many of the correct tiles, but leave clear gaps in the generated path (Fig.~\ref{fig:pf_mlp}). Instead of learning a localized, convolution-type operation at each layer, the MLP may be behaving more like an auto-encoder (owing to their repeated encoder/decoder block architecture), memorizing the label relative to the entire maze and reproducing it with relatively high accuracy but without preserving local coherence.

NCAs fail to perfectly generalize the diameter problem. They sometimes select the wrong branch toward the end of a largely correct path (Fig.~\ref{fig:diam_fork}), which is a less crucial mistake. 
If there are two equivalent diameters, the model will sometimes activate both sub-paths. 

\section{Conclusion}

In this paper, we introduce neural implementations of the shortest path finding and diameter problems. 
We posit that these hand-coded models can provide insight into learning more general models in more complex pathfinding-related tasks.
We validate our claims by showing that architectural modifications inspired by hand-coded solutions lead to models that generalize better.

One limitation of our method is that it deals only with mazes defined on a grid.
While our neural BFS model could be readily adapted to arbitrary graphs (i.e. translated from an NCA to a GCN architecture), our DFS implementation relies on the convolutional structure of NCAs.
One could imagine re-implementing the sequential queuing logic of our DFS-NCA in a Graph-NCA, by using the message-passing layer of an anisotropic GNN such as a GAT.
Such hand-coded solutions could be key to understanding how to scale the strong generalization ability exhibited by learned NCAs on complex grid mazes to similarly complex mazes on arbitrary graphs.



Future work may also benefit from these differential sub-modules directly, either using them to augment a learning model or using them as an adaptable starting point to further improve existing algorithms or fit them to a particular context.


\bibliographystyle{unsrtnat}
\bibliography{refs}

\newpage





\appendix

\section{Appendix}
\label{s:appendix}

\subsection{Qualitative Analysis}
\begin{figure}[ht]
\begin{subfigure}{1.0\linewidth}
\includegraphics[width=2.7cm,trim={0 0 19 0},clip]{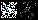}
\includegraphics[width=2.7cm,trim={19 0 0 0},clip]{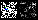}
\includegraphics[width=2.7cm,trim={19 0 0 0},clip]{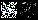}
\includegraphics[width=2.7cm,trim={19 0 0 0},clip]{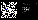}
\includegraphics[width=2.7cm,trim={19 0 0 0},clip]{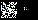}
\caption{\textbf{MLP shortest-path behavior.} The model reconstructs a small patch approximating the target path, but misses and misplaces a few tiles, and fails to maintain local coherence.}
\label{fig:pf_mlp}
\end{subfigure}
\begin{subfigure}{1.0\linewidth}
\includegraphics[width=2.7cm,trim={0 0 19 0},clip]{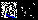}
\includegraphics[width=2.7cm,trim={19 0 0 0},clip]{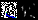}    
\includegraphics[width=2.7cm,trim={19 0 0 0},clip]{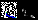}
\includegraphics[width=2.7cm,trim={19 0 0 0},clip]{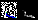}
\includegraphics[width=2.7cm,trim={19 0 0 0},clip]{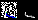}
\caption{\textbf{GCN shortest-path behavior.} The model produces a (somewhat blurry) reconstruction of the optimal path. It has difficulty reconstructing longer paths from the training set.}
\label{fig:pf_gcn}
\end{subfigure}
\begin{subfigure}{1.0\linewidth}
\includegraphics[width=2.7cm,trim={0 0 19 0},clip]{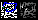}
\includegraphics[width=2.7cm,trim={19 0 0 0},clip]{Figures/pf_16/NCA/render_3.png}
\includegraphics[width=2.7cm,trim={19 0 0 0},clip]{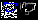}
\includegraphics[width=2.7cm,trim={19 0 0 0},clip]{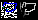}
\includegraphics[width=2.7cm,trim={19 0 0 0},clip]{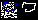}
\caption{\textbf{NCA shortest-path behavior.} The model appears to propagate activation out from both source and target, which meet in the middle of the optimal path and proceed to reinforce it.}
\label{fig:nca_pf_succ}
\begin{subfigure}{1.0\linewidth}
\includegraphics[width=2.7cm,trim={0 0 35 0},clip]{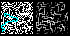}
\includegraphics[width=2.7cm,trim={35 0 0 0},clip]{Figures/diameter_32/NCA_succ/render_900.png}
\includegraphics[width=2.7cm,trim={35 0 0 0},clip]{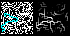} 
\includegraphics[width=2.7cm,trim={35 0 0 0},clip]{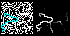} 
\includegraphics[width=2.7cm,trim={35 0 0 0},clip]{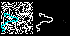}
\caption{\textbf{NCA diameter behavior.} The model successfully identifies the diameter of the maze.}
\label{fig:nca_diam_succ}
\end{subfigure}
\end{subfigure}
\begin{subfigure}{1.0\linewidth}
\includegraphics[width=2.7cm,trim={0 0 35 0},clip]{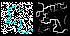}
\includegraphics[width=2.7cm,trim={35 0 0 0},clip]{Figures/diameter_32/bad_fork/render_777.png}    
\includegraphics[width=2.7cm,trim={35 0 0 0},clip]{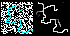}    
\includegraphics[width=2.7cm,trim={35 0 0 0},clip]{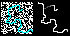}    
\includegraphics[width=2.7cm,trim={35 0 0 0},clip]{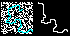}    
\caption{\textbf{NCA diameter behavior.} The model selects the incorrect (shorter) upper fork.}
\label{fig:diam_fork}
\end{subfigure}
\caption{Behavior of various models on shortest-path/diameter path-finding problems. GCN and MLP models are seen to fail frequently on simpler mazes from the training set for the shortest-path problem. NCAs can master the most complex mazes from the training set while sometimes generalizing to larger test mazes, on both the shortest-path and (more challenging) diameter problem.}
\label{fig:appendix_mazes}
\end{figure}

\subsection{Hand-coded NCA}
\label{s:handcodeappendix}

In discussion of the hand-coded implementations, we denote $W^{(m,n)}_k$ as the $(m,n)^{\rm th}$ element of the weight matrix $\*W_k$.
We will define a number of elementary convolutional weight matrices, then specify how they are combined in a single convolutional layer to propagate activation over various channels.
Finally, we will specify activation functions and skip connections applied to each forward pass through this convolutional layer.
Initially, the hidden activation is all-zeros.
We assume that, prior to each pass through the NCA, the hidden activation is concatenated channel-wise with a one-hot encoding of the input maze.

\subsubsection{Dijkstra map generation}
\label{s:handcodeappendix_dijkstra}

In Dijkstra map generation, we define three channels: ${\rm flood}_s$, ${\rm flood}_t$ and ${\rm age}$. 

\paragraph{Convolutional weights.}

We define an identity-preserving convolutional weight matrix $\*W_1\in \R^{3\times 3}$ with $\*W_1^{(1,1)}=1$; otherwise, $\*W_1^{(k,l)}=0$ so only center tile of the weight matrix is active. This matrix is used to keep a channel active once it is activated (e.g., once a tile is flooded, it needs to stay flooded). 
It is also used to produce the initial activation in the $\mathrm{flood}$ channels from $\mathrm{source}$ and $\mathrm{target}$ channels, and prevent tiles with active $\mathrm{wall}$ channels (i.e. walls in the maze) from being flooded. 

Then, we define $\*W_T\in \R^{3\times 3}$, which is the Von Neumann neighborhood, with $\*W_T^{(k,l)}=1$ if $(k,l) \in \{(0,1), (1,0), (1,1), (1,2), (2,1)\}$; otherwise, $\*W_T^{(k,l)}=0$ and it's used to flood the tiles next to flood since one can only move in this four directions. 

\begin{align*}
\*W_1 \coloneqq \begin{bmatrix}
0 & 0 & 0\\
0 & 1 & 0\\
0 & 0 & 0
\end{bmatrix} &&
\*W_T \coloneqq \begin{bmatrix}
0 & 1 & 0\\
1 & 1 & 1\\
0 & 1 & 0
\end{bmatrix}
\end{align*}

In Table~\ref{tab:dijk}, we use the elementary weight matrices $\*W_1$ and $\*W_T$ to define the weight matrices between channels in the convolutional layer of which our hand-coded Dijkstra activation map-generating NCA consists.

\begin{table}[hbp]
\centering
\begin{tabular}{|c|c|r|}
\cline{1-3}
Convolutional slice & Channel relationship                  & Weight matrix   \\ \hline
$\*W_{\mathrm{BFS}}[\mathrm{flood}\,, \mathrm{source}]$ & \multicolumn{1}{|r|}{source $\xrightarrow[]{}$ ${\rm flood}_{s}$} & \begin{tabular}[c]{@{}l@{}} $\*W_1$\end{tabular} \\ \hline
. & \multicolumn{1}{|r|}{${\rm flood}_{s}$ $\xrightarrow[]{}$ ${\rm flood}_{s}$} & \begin{tabular}[c]{@{}l@{}} $\*W_{T}$ \end{tabular} \\ \hline
. & \multicolumn{1}{|r|}{wall $\xrightarrow[]{}$ ${\rm flood}_s$}  & $-6 \*W_{1}$                                                                           \\ \hline
 . & \multicolumn{1}{|r|}{target $\xrightarrow[]{}$ ${\rm flood}_{t}$} & \begin{tabular}[c]{@{}l@{}} $\*W_1$ \end{tabular} \\ \hline
  & \multicolumn{1}{|r|}{${\rm flood}_{t}$ $\xrightarrow[]{}$ ${\rm flood}_{t}$} & \begin{tabular}[c]{@{}l@{}} $\*W_T$ \end{tabular} \\ \hline
   & \multicolumn{1}{|r|}{wall $\xrightarrow[]{}$ ${\rm flood}_t$}  & $-6 \* W_1$                                                                         \\ \hline
  & \multicolumn{1}{|r|}{${\rm flood}_s$ $\xrightarrow[]{}$ age}   & $\*W_1$                                                                            \\ \hline
 & \multicolumn{1}{|r|}{${\rm flood}_t$ $\xrightarrow[]{}$ age}   & $\*W_1$                                                                            \\ \hline
 
 & \multicolumn{1}{|r|}{age $\xrightarrow[]{}$ age}     & $\*W_1$                                                                            \\ \hline
\end{tabular}
\caption{ hand-coded weights between channels of NCA for \textbf{Dijkstra map generation}.  }
\label{tab:dijk}
\end{table}


The overall convolutional layer $\*W_{\mathrm{BFS}} \in \R^{3\times 7\times 3 \times 3}$ consists of the channel-to-channel kernels given in Table~\ref{tab:dijk}. 
For each $3\times 3$ patch of the (padded) input activation, it maps to the next value at the center cell; it maps from the $3$ hidden activation channels ($\{\mathrm{age}, \mathrm{flood}_s, \mathrm{flood}_t\}$) plus the $4$ channels required for one-hot encoding the maze $\{\mathrm{empty}, \mathrm{wall}, \mathrm{source}, \mathrm{target}\}$---which is concatenated with the input activation at each step---to the $3$ hidden channels at the next step.

The floods $\mathrm{flood}_s$ and $\mathrm{flood}_t$ flow from tiles with active $\mathrm{source}$ and $\mathrm{target}$ tiles, respectively (i.e. sources/targets in the input maze), to empty tiles, stopping at walls. While the $\mathrm{age}$ channel will first activate with the appearance of a $\mathrm{flood}$ channel, then increment at each following timestep. Once the separate floods meet, the activation map $\*X^{\mathrm{age}}$ can then be used to reconstruct optimal paths (which process is detailed in the path extraction NCA in the following section).

\paragraph{Forward pass.}

At each step through the hand-coded BFS-NCA, we apply the convolutional weights to the input activation, then apply a step function to the $\mathrm{flood}$ channel, so that its output lies between $0$ and $1$. (Since the $\mathrm{flood}$ activation is always integer-valued before being input to the activation function, this could also be achieved with duplicate channels, using ReLU's, with biases to off-set them, effectively resulting in a step function that is linear for $x\in [0,1]$. For simplicity, we simply apply a step function, but note that a ReLU network could represent the same algorithm.)

\begin{equation*}
\mathrm{step} = 
\begin{cases}
    0 & \text{for } x < 0 \\
    1 & \text{for } x \geq 0
\end{cases}
\end{equation*}

Let $\*X_t$ denote the hidden activation at time-step $t$, with $\*X_t \in \R^{3\times w \times h}$ (where $w$ and $h$ are the width and height of the input maze), and $\*X_t = \*0$ when $t=0$; and let $X_{\mathrm{maze}}$ denote the one-hot encoding of the input maze, with $\*X_{\mathrm{maze}} \in \R^{4\times w \times h}$.
We denote by $\*X^{i, j}$ the slice of the tensor $\*X$ taken at spatial co-ordinates $i, j$ (where these spatial coordinates correspond to the last two dimensions of the tensor), and by $\*X^\mathrm{channel}$ the slice of the tensor $\*X$ corresponding to the channel $\mathrm{channel}$ (where channels lie along the first dimension of the tensor).
The $\mathrm{concatenate}$ operation acts along the channel dimension.
To denote setting the value of an activation $\*X$ at channel $\mathrm{chan}$, we write $\*X^{\mathrm{chan}} \leftarrow \cdots$; generally, this operation is individually applied to each $(x, y)$ tile of $\*X$.
At time-step $t$ the forward pass operates as follows:

\begin{align*}
    \*X_t & \xleftarrow[]{} \mathrm{concatenate}\left(\*X_t, \*X_{\mathrm{maze}}\right)\\
    \forall x, y, \, \*X^{x, y}_{t+1} & \xleftarrow[]{} \sum_{i=0}^2\sum_{j=0}^2 \*W^{i, j}_{\mathrm{BFS}} \odot \*X_t^{x+i-1, y+j-1}\\
    \*X^{\mathrm{flood}}_{t+1} & \xleftarrow{} \mathrm{step}\left(\*X^\mathrm{flood}_{t+1}, 0, 1\right)
\end{align*}

The BFS-NCA will terminate when the source and target floods overlap on some tile.

\subsubsection{Path Extraction}
\label{s:handcodeappendix_path_extraction}

We now construct an NCA for path extraction (PE-NCA) which will take the output of the BFS-NCA and additionally perform path extraction. Path  extraction starts once source and target floods, which are defined in $\mathrm{Dijkstra}$ collides. We use the $\mathrm{age}$ activations from $\mathrm{Dijkstra}$ output to reconstruct the optimal path(s) between the corresponding source and target. 

\paragraph{Convolutional weights.}
For path extraction, we add $5$ new channels: $\mathrm{path}$ channel, which will ultimately output the binary map of the optimal path(s), and 4 directional path-activation channels (for detecting the presence of a path coming from the right, left, top, and bottom neighbors) denoted as $\mathrm{path}_{i,j}$ for $(i,j) \in \{(0,1), (1,0),  (1,2), (2,1) \}$.  Then, we define weight matrices for detecting adjacent activations: $\*W_{(i,j)} \in \R^{3\times3} $, with $W^{(k,l)}_{(i,j)}\coloneqq 1$ if $(k,l) = (i,j)$, and $W^{(k,l)}_{(i,j)} \coloneqq 0$ otherwise. 

\begin{align*}
\*W_{1,0} \coloneqq \begin{bmatrix}
0 & 0 & 0\\
1 & 0 & 0\\
0 & 0 & 0
\end{bmatrix} &&
\*W_{0,1} \coloneqq \begin{bmatrix}
0 & 1 & 0\\
0 & 0 & 0\\
0 & 0 & 0
\end{bmatrix} &&
\*W_{1,2} \coloneqq \begin{bmatrix}
0 & 0 & 0\\
0 & 0 & 1\\
0 & 0 & 0
\end{bmatrix} &&
\*W_{2,1} \coloneqq \begin{bmatrix}
0 & 0 & 0\\
0 & 0 & 0\\
0 & 1 & 0
\end{bmatrix}
\end{align*}

\begin{table}
\centering
\begin{tabular}{|l|r|r|}
\cline{1-3}
Channel relationship                                                      & Weight matrix                  & Bias                               \\ \hline
\multicolumn{1}{|r|}{\begin{tabular}[r]{@{}l@{}} ${\rm flood}_s \xrightarrow[]{} \mathrm{path}$ \end{tabular}}  & $\*W_1$ &   \\ \hline
\multicolumn{1}{|r|}{\begin{tabular}[r]{@{}l@{}} ${\rm flood}_t \xrightarrow[]{} \mathrm{path}$ \end{tabular}}  & $\*W_1$ &   \\ \hline
\multicolumn{1}{|r|}{\begin{tabular}[r]{@{}l@{}} $\mathrm{path}$ \end{tabular}}  &  & -1 \\ \hline
\multicolumn{1}{|r|}{\begin{tabular}[r]{@{}l@{}}age $\xrightarrow[]{}$ ${\rm path}_{i,j}$ \end{tabular}}  & $2( \*W_{(i,j)} - \*W_1$) & \\\hline
\multicolumn{1}{|r|}{\begin{tabular}[r]{@{}l@{}}path $\xrightarrow[]{}$  ${\rm path}_{i,j}$ \end{tabular}} & $\*W_{ (i,j)}$&  \\ \hline
\end{tabular}
\caption{hand-coded weights between channels of NCA for \textbf{path extraction}. }
\label{tab:pathextract}
\end{table}

We can then construct the overall convolutional path extraction matrix $\*W_{\mathrm{PE}}$, with $\*W_{\mathrm{PE}} \in \R^{5 \times 8 \times 3\times 3}$ using the channel-to-channel weight matrices detailed in Table~\ref{tab:pathextract} (and setting all other weights to $0$).
For each $3\times 3$ patch of the (padded) input activation, it maps to the next value at the center cell; it maps from its own $5$ hidden activation channels (the $4$ $\mathrm{path}_{i,j}$ activations, and the overall $\mathrm{path}$ activation) plus the $3$ hidden channels resulting from the BFS-NCA---whose output is concatenated with the input activation at each step---to the $5$ hidden channels at the next step.

The identity weights from $\mathrm{flood}_s$ and $\mathrm{flood}_t$ to $\mathrm{path}$, combined with the bias of $-1$ applied to $\mathrm{path}$, ensure that path activations will first appear on any tiles where these floods have overlapped (seeing as these tiles will correspond to mid-points in the optimal path(s)).
The weights from $\mathrm{age}$ and $\mathrm{path}$ to each $\mathrm{path_{i,j}}$ are chosen so that when a path activation should ``flow'' to an adjacent tile in a given direction---which is the case exactly when this neighbor's age is greater than the current cell's by $1$---the corresponding $\mathrm{path}_{i,j}$ activation will be exactly $-1$.

\paragraph{Forward pass.}

At each step through the PE-NCA, we apply the convolutional weights and biases corresponding to path extraction---$\*W_{\mathrm{PE}}$ and $\*b_{\mathrm{PE}}$, respectively.
Then, we apply a saw-tooth activation function to the directional path-activation channels. (Similar to the $\mathrm{step}$ function in the forward pass of the BFS-NCA, this $\mathrm{sawtooth}$ activation could be replicated with 3 $\ReLU$'s, and corresponding additional channels.)

\begin{align*}
    \mathrm{sawtooth}_{a}(x) & = 
    \begin{cases} 
        0 & \text{for } x < -2 \\
        x + a - 1 & \text{for } x\in [a - 1, a] \\
        -x + a + 1 & \text{for } x\in [a, a + 1] \\
        1 & \text{for } x > a + 1 
    \end{cases}
\end{align*}

Since the directional path-activations are always integer-values, we can apply $\mathrm{sawtooth}_{-1}$ to them in order to obtain activations of $1$ wherever they have value $-1$ (i.e., they should accept an adjacent path activation and are part of the optimal path by virtue of their age-difference with their path-activated neighbor), and $0$ everywhere else.

Then, we set the path channel to be $1$ if any of the directional path activations at the corresponding tile are equal to $1$.
This can be achieved by taking the sum of the directional path activations (which are either $0$ or $1$ after the sawtooth activation), and applying a step function.

\begin{align*}
    \*X_t & \xleftarrow[]{} \mathrm{concatenate}\left(\*X_t, \*X_{\mathrm{BFS}}\right)\\
    \forall x, y, \, \*X^{x, y}_{t+1} & \xleftarrow[]{} \sum_{i=0}^2\sum_{j=0}^2 \*W^{i, j}_{\mathrm{PE}} \odot \*X_t^{x+i-1, y+j-1} + \*b_{\mathrm{PE}}\\
    \forall i, j,\,\*X^{\mathrm{path}_{i,j}}_{t+1} & \xleftarrow{} \mathrm{sawtooth}_{-1}\left(\*X^{\mathrm{path}_{i,j}}_{t+1}\right)\\
    \*X^{\mathrm{path}}_{t+1} & \xleftarrow{} \mathrm{step}\left(\sum_{(i, j)} \*X^{\mathrm{path}_{i, j}}\right)
\end{align*}

\begin{figure*}[t]
\begin{subfigure}{\textwidth}
    \centering
    \includegraphics[width= \linewidth]{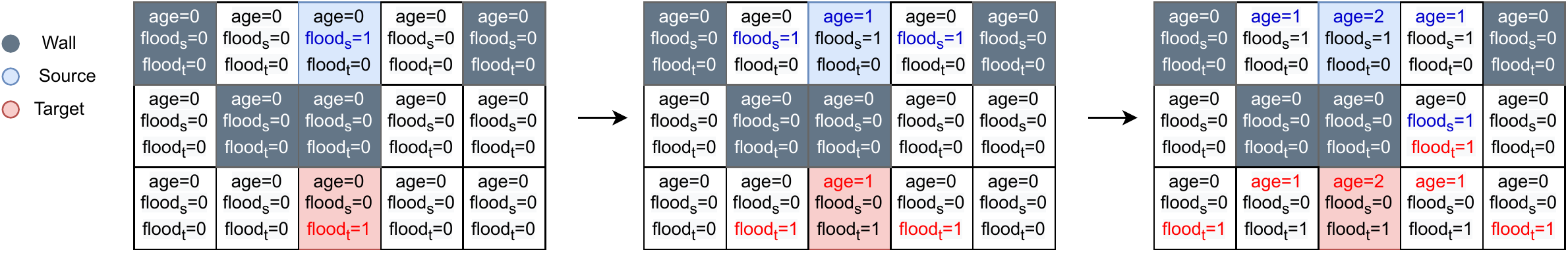}
    \caption{Demonstration of hand-coded Dijkstra map generation by flooding from both source and target. The blue color age and floor activation are due to source and red one is due to target. The map generation is completed when both flood channels are active. }
    \label{fig:conv}
\end{subfigure}
\begin{subfigure}{\textwidth}
    \centering
    \includegraphics[width= \linewidth]{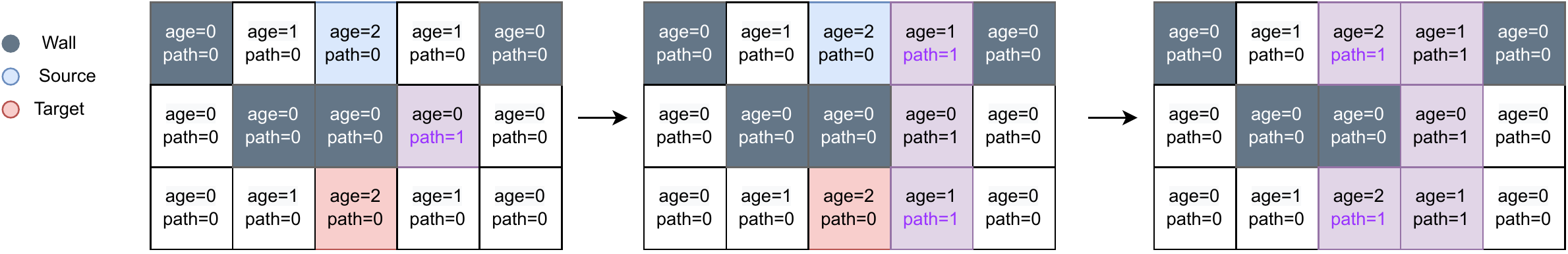}
    \caption{Demonstration of hand-coded path-extraction by using Dijkstra map. The first activation of the path is where source and target flood meets. The purple color demonstrates the current path activation tile. The path extraction is completed when path channel of source and target is activated. }
    \label{fig:extract}
\end{subfigure}
\caption{Demonstration of hand-coded NCA for a) Dijkstra map generation, b) path extraction}
\label{fig:hand-code}
\end{figure*}

We illustrate the step-by-step operation of the path extraction NCA in Figure~\ref{fig:extract}.

\subsubsection{DFS}
\label{s:handcodeappendix_dfs}

\paragraph{Convolutional weights.}
Let us again denote an identity-preserving convolutional weight matrix $\*W_2 \in \R^{5 \times 5}$ with $W^{(2,2)}_2=1$ (at the center of the weight matrix) and $W^{(k,l)}_2=0$ everywhere else. 
$\*W_2$ is similar to $\*W_1$, in that only the center tile is active, allowing for the transfer of activations between channels, the inhibition of one activation by another (i.e. having $\mathrm{wall}$ activations prevent the activation of $\mathrm{route}$ or $\mathrm{stack}$ channels), and the holding-constant of activations across time-steps. 

Then, we define the adjacent-activation weight matrices $\*W_{5,(i,j)}\in \R^{5 \times 5}$ for $(i,j) \in \{(1,2), (2,1), (3,2), (2,3)\}$, which are used to check whether neighbors have any adjacent $\mathrm{route}$ activations, with $W^{(k,l)}_{5,(i,j)} = 1$ for $(k,l)= (i,j)$, and $W^{(k,l)}_{5,(i,j)} =0$ otherwise. 
Just as the identity-preserving matrix $\*W_2$ is equivalent to $\*W_1$, plus a border of 0-padding, so is each matrix $\*W_{5,(i,j)}$ to the corresponding $3\times 3$ adjacent-activation matrix $\*W_{(i,j)}$.

In DFS, we propagate the $\mathrm{route}$ activation (equivalent to $\mathrm{flood}$ in BFS) sequentially. To this end, we combine the adjacent-activation matrices above with priority weight-matrices. These priority weights are the only matrices to make use of our $5\times 5$ kernel; their job is to detect whether---given that some neighbor of a given cell has an active route which should be propagated to said cell---there is not some neighbor's neighbor to whom this activation should be propagated first.

For moving down, the priority weight matrix is defined as $\*W_{p, (1,2)} \in \R^{5 \times 5}$ and it is equal to $\*0$ since moving down is the highest-priority move. 
Then, in order to check the priority of moving right, we define $\*W_{p, (2,1)} \in \R^{5 \times 5}$, with $W^{3,1}_{p, (2,1)} = 1$; and $0$ everywhere else. 
This matrix effectively checks whether the neighbor to our left has some neighbor below it to whom it could pass its `route' activation, in which case this cell should take priority over us.
Similarly, we define $\*W_{p, (3,2)} \in \R^{5 \times 5}$ for the upward move priority, with $W^{m,n}_u\coloneqq1$ if $(m,n) \in \{(4,2), (3,3)\}$, and $0$ otherwise. 
This matrix is used to check whether the neighbor above us as a neighbor either below or to its right to whom it should pass activation first.
Finally, moving left is the least prioritized action, and is represented with $\*W_{p, (2,3)} \in \R^{5 \times 5}$ , where $W^{(m,n)}_{p, (2,3)}=1$ if $(m,n) \in \{(1,3), (2,4), (3,3)\}$, and $0$ otherwise.
This matrix only sends activation from the neighbor to a cell's right if this neighbor has no cells below, to the right, or above it, which would then take priority over us.

\begin{align*}
\*W_{p,(2,1)} \coloneqq \begin{bmatrix}
0 & 0 & 0 & 0 & 0\\
0 & 0 & 0 & 0 & 0\\
0 & 0 & 0 & 0 & 0\\
0 & 1 & 0 & 0 & 0\\
0 & 0 & 0 & 0 & 0
\end{bmatrix} &&
\*W_{p,(3,2)} \coloneqq \begin{bmatrix}
0 & 0 & 0 & 0 & 0\\
0 & 0 & 0 & 0 & 0\\
0 & 0 & 0 & 0 & 0\\
0 & 0 & 0 & 1 & 0\\
0 & 0 & 1 & 0 & 0
\end{bmatrix} &&
\*W_{p,(2,3)} \coloneqq \begin{bmatrix}
0 & 0 & 0 & 0 & 0\\
0 & 0 & 0 & 1 & 0\\
0 & 0 & 0 & 0 & 1\\
0 & 0 & 0 & 1 & 0\\
0 & 0 & 0 & 0 & 0
\end{bmatrix} 
\end{align*}

When the route activation is \textit{not} propagated to some adjacent, available neighbor as a result of one of the priority-rankings as defined above, we add a binary $\mathrm{stack}$ activation to the ignored tile.
We then count the age of this $\mathrm{stack}$ activation with the $\mathrm{stack\_rank}$ channel.
When the pebble becomes stuck, the least ${\rm stack\_rank}$ activations will correspond to the edges that the pebble has ignored most recently.
Since, if the pebble ignores multiple edges at a given iteration (e.g., by moving downward when both right and left neighbors are also available), there will be multiple equivalent $\rm stack\_rank$ activations, we further encode their directional priority on the stack via the $\rm stack\_direction$ channel.

For the priority of the direction, we define $\*W_p \in \R^{5\times 5}$ and $W^{1,2}_p= 0.2$, $W^{2,1}_p= 0.4$, $W^{3,2}_p= 0.6$, $W^{2,3}_p= 0.8$; otherwise, $W^{k,l}_p= 0$. Then, we define $\*W_{a} \in \R^{5\times 5}$ to check the neighbor tiles to follow the `pebble' and $W^{(m,n)}_a =1$ if $(m,n) \in \{(1,2), (2,1), (3,2), (2,3)\}$; otherwise, $W^{(m,n)}_a =0$.  

\begin{align*}
\*W_p \coloneqq \begin{bmatrix}
0 & 0 & 0 & 0 & 0\\
0 & 0 & \frac{1}{5} & 0 & 0\\
0 & \frac{4}{5} & 0 & \frac{2}{5} & 0\\
0 & 0 & \frac{3}{5} & 0 & 0\\
0 & 0 & 0 & 0 & 0
\end{bmatrix} &&
\*W_a \coloneqq \begin{bmatrix}
0 & 0 & 0 & 0 & 0\\
0 & 0 & 1 & 0 & 0\\
0 & 1 & 0 & 1 & 0\\
0 & 0 & 1 & 0 & 0\\
0 & 0 & 0 & 0 & 0
\end{bmatrix} 
\end{align*}

Note that in the event a cell which is already ``on the stack'' is added to it again (i.e., the pebble passes by it at two neighboring cells, without having moved onto it), the $\rm stack\_rank$ and $\rm stack\_direction$ activations at this tile are effectively reset or overwritten, to correspond to this more ``recent'' edge in terms of the pebble's traversal over the graph.

Having defined these elementary weights, we give the full specification of the hand-coded DFS-NCA in Table~\ref{tab:dfs}.
The result is a convolutional weight matrix $\*W_{\mathrm{DFS}}\in \R^{3\times 3}$

\begin{table}
\centering
\parbox{.45\linewidth}{
\begin{tabular}{|c|r|}
\cline{1-2}
Chanel relationship                                                      & Weight   \\ \hline
\multicolumn{1}{|r|}{$\mathrm{source} \xrightarrow[]{} \mathrm{route}$}                   & $\*W_2$           \\ \hline
\multicolumn{1}{|r|}{$\mathrm{route}$ $\xrightarrow[]{}$  $\mathrm{route}$}                    & $\*W_2$           \\ \hline
\multicolumn{1}{|r|}{$\mathrm{wall}$ $\xrightarrow[]{}$  $\mathrm{route}$}                     & $-  \*W_2$  \\ \hline
\multicolumn{1}{|r|}{$\mathrm{stack}$ $\xrightarrow[]{}$ $\mathrm{route}$}                      & $-  \*W_2$  \\ \hline
\multicolumn{1}{|r|}{$\mathrm{route} \xrightarrow[]{} \mathrm{route}_{i,j}$}            & $\*W_{5, (i,j)}$ \\ \hline
\multicolumn{1}{|r|}{$\mathrm{route} \xrightarrow[]{} \mathrm{route}_{i,j}$}            & $\*W_{p, (i,j)}$ \\ \hline
\multicolumn{1}{|r|}{$\mathrm{empty} \xrightarrow[]{} \mathrm{route}_{i,j}$}            & $-\*W_{p, (i,j)}$ \\ \hline
\multicolumn{1}{|r|}{$\mathrm{source} \xrightarrow[]{} \mathrm{route}_{i,j}$}            & $-\*W_{p, (i,j)}$ \\ \hline
\multicolumn{1}{|r|}{$\mathrm{target} \xrightarrow[]{} \mathrm{route}_{i,j}$}            & $-\*W_{p, (i,j)}$ \\ \hline
\multicolumn{1}{|r|}{$\mathrm{pebble}$ $\xrightarrow[]{}$ $\mathrm{stack}$}                    & $\*W_a$           \\ \hline
\multicolumn{1}{|r|}{$\mathrm{stack}$  $\xrightarrow[]{}$ $\mathrm{stack}$}                    & $\*W_2$           \\ \hline
\multicolumn{1}{|r|}{$\mathrm{wall}$ $\xrightarrow[]{}$ $\mathrm{stack}$}                      & $-2  \*W_2$  \\ \hline
\multicolumn{1}{|r|}{$\mathrm{route}$ $\xrightarrow[]{}$ $\mathrm{stack}$}                     & $-2  \*W_2$  \\ \hline
\multicolumn{1}{|r|}{$\mathrm{pebble}\xrightarrow[]{} \rm stack\_direction$}          & $\*W_p$           \\ \hline
\multicolumn{1}{|r|}{$\mathrm{stack\_direction} \xrightarrow[]{} \mathrm{stack\_direction}$} & $\*W_2$           \\ \hline
\multicolumn{1}{|r|}{$\mathrm{wall} \xrightarrow[]{} \mathrm{stack\_direction}$}            & $-2\*W_2$  \\ \hline
\multicolumn{1}{|r|}{$\mathrm{route} \xrightarrow[]{} \mathrm{stack\_direction}$}       & $-2\*W_2$  \\ \hline
\multicolumn{1}{|r|}{$\mathrm{pebble}\xrightarrow[]{} \rm stack\_rank$}          & $\*W_2$           \\ \hline
\multicolumn{1}{|r|}{$\mathrm{stack\_direction} \xrightarrow[]{} \mathrm{stack\_rank}$} & $\*W_2$           \\ \hline
\multicolumn{1}{|r|}{$\mathrm{wall} \xrightarrow[]{} \mathrm{stack\_rank}$}            & $-2\*W_2$  \\ \hline
\multicolumn{1}{|r|}{$\mathrm{route} \xrightarrow[]{} \mathrm{stack\_rank}$}       & $-2\*W_2$  \\ \hline
\multicolumn{1}{|r|}{$\mathrm{stack} \xrightarrow[]{} \mathrm{stack\_rank}$}                & $ \*W_2$          \\ \hline
\multicolumn{1}{|r|}{$\mathrm{pebble} \xrightarrow[]{} \mathrm{since\_binary}$}               & $ \*W_2$          \\ \hline
\multicolumn{1}{|r|}{$\mathrm{since\_binary} \xrightarrow[]{} \mathrm{since\_binary}$}               & $ \*W_1$          \\ \hline

\multicolumn{1}{|r|}{$\mathrm{since\_binary} \xrightarrow[]{} \mathrm{since\_binary}$}                & $ \*W_1$          \\ \hline
\multicolumn{1}{|r|}{$\mathrm{since} \xrightarrow[]{} \mathrm{since}$}                & $ \*W_1$          \\ \hline
\end{tabular}
\caption{hand-coded weights between channels of NCA for \textbf{DFS}.}
\label{tab:dfs}
}
\end{table}

The $\mathrm{source}$ tile will result in the initial $\mathrm{route}$ activation.
Once a tile has an active $\mathrm{route}$, it will be maintained at following timesteps.
$\mathrm{wall}$ and $\mathrm{stack}$ activations will inhibit the appearance of a $\mathrm{route}$ activation.
Directional $\mathrm{route}$ activations will appear if the tile is empty and if any of the higher priority neighbors in $\*W_{p,(i,j)}$ are available (that is, they are $\mathrm{empty}$, $\mathrm{source}$ or $\mathrm{target}$ tiles, without any existing $\mathrm{route}$ activation already present on them).

If the $\mathrm{pebble}$ passes by a some tile at any of its neighbors (given by $\*W_a$), this tile is added to the $\mathrm{stack}$, with the $\mathrm{stack}$ activation sustaining itself across following timesteps.
This $\mathrm{stack}$ activation is nullified, however, if there is a $\mathrm{wall}$ on the current tile, or is received $\mathrm{route}$ activation at the next iteration.
(A $\mathrm{stack}$ activation thus only becomes active when the $\mathrm{pebble}$ ``ignores'' an available tile as a result of directional priority.)

The $\mathrm{stack\_direction}$ channel is similarly self-sustaining, and is activated when the $\mathrm{pebble}$ passes by the current tile without having been inhibited by a $\mathrm{wall}$ and without moving onto the tile at the next iteration, though this time it is activated according to the directional priority matrix $\*W_p$.

The $\mathrm{stack\_direction}$ channel is, again, self-sustaining and activated when ignored by a pebble. Additionally, it increments at each following time-step as long as a $\mathrm{stack}$ remains present at the current tile.

\paragraph{Forward pass.}

We apply a step function to the directional $\mathrm{route}_{i,j}$ activation channels.
The result will indicate whether, given the presence of adjacent $\mathrm{route}$ activations, the priority allotted to any relevant and available second-neighbors, and the presence of any $\mathrm{wall}$ or $\mathrm{stack}$ activations, $\mathrm{route}$ activation if flowing into a given tile in a given direction.
As a first step toward, determining the $\mathrm{route}$ activation at a given tile, we take the sum of the directional route-activations at this tile, and apply a step function to the result.
This will indicate whether $\mathrm{route}$ activation is flowing into this tile from \textit{any} direction.
We then add the result to the $\mathrm{route}$ activation and take the step function again: this will ensure that any blocking $\mathrm{wall}$ or $\mathrm{stack}$ activations prevent the final $\mathrm{route}$ activation.

We apply ReLU functions to the stack, stack rank, and stack direction channels.
To address the event of a given tile being re-added to the stack before being popped from it, we apply a sawtooth to detect tiles with a $\mathrm{stack}$ activation equal to $2$, resulting in a binary array, $\mathbf{dbl\_stacked} \in \{0,1\}^{w\times h}$.
To overwrite the old $\mathrm{stack\_rank}$ and $\mathrm{stack\_direction}$ activations (and reset the $\mathrm{stack}$ activation to $1$), we subtract from each of these channels the product of their value at the preceding time-step ($t-1$) with $\mathbf{dbl\_stacked}$.
For the $\mathrm{stack\_rank}$ channel, we must subtract its previous value $+ 1$, since its $\mathrm{stack\_rank}$ has incremented during the most recent convolution (as occurs at each time-step) in addition to having increased as a result of its new position at the top of the stack.

The $\mathrm{pebble}$ channel is given by the difference between the $\mathrm{route}$ activations at the current time-step and those at the previous time-step, so that there is at most a single $\mathrm{pebble}$ on the board, in the place of the single newly-added $\mathrm{route}$ activation.
If the pebble is stuck, there is no pebble activation, so we take the spatial max-pool over the pebble channel and apply a $\mathrm{sawtooth_{0}}$ to obtain a binary value corresponding to the pebble's being stuck.

In case the pebble is stuck, we need to know which tile on the stack to pop next, taking into account both the directional priority and time spent on the stack of each tile.
We thus calculate the $\mathbf{total\_rank}$ of each tile (stored as a temporary variable, with $\mathbf{total\_rank}\in \R^{w\times h}$) by taking the sum of its $\mathrm{stack\_rank}$ and $\mathrm{stack\_direction}$ channels.
The tile with least non-zero $\mathrm{total\_rank}$ is the next tile to be popped.
We first replace all $0$ activations in the $\mathrm{total\_rank}$ array by adding to this array the result of $\mathrm{step}_0(\mathbf{total\_rank}$ multiplied by a large integer $L$ (i.e. the maximum value of $\mathrm{stack\_rank} + 1$
Then, we take the minimum over the resulting $\mathbf{total\_rank}$: this is the $\mathrm{min\_total\_rank} \in \N_0$ (the same could be achieved by flipping signs and using a max-pooling layer).
(Note that if $\mathrm{min\_total\_rank} = L$, then the stack is empty, we are effectively done, and the following steps will have no effect.)

Wherever the difference $\mathrm{total\_rank} - \mathrm{min\_rank}$ is $0$, we must pop from the stack, provided also that the pebble is stuck.
Thus, to obtain a binary array $\mathbf{is\_popped} \in \{0,1\}^{w\times h}$ with a $1$ in at the one tile to be popped (if any), we take the product of $\mathrm{sawtooth}_{0}$ applied to the above difference, with the binary variable indicating the pebble's being stuck.
Finally, to release the given tile from the stack, we subtract from the $\mathrm{stack}$, $\mathrm{stack\_rank}$ and $\mathrm{stack\_direction}$ channels their product with $\mathrm{is\_popped}$, zeroing-out these channels and dis-inhibiting the flow of $\mathrm{route}$ activation to these tiles at following time-steps.

We note that $\mathrm{since}$ channel activation happens when $\mathrm{pebble}$ triggers the $\mathrm{stack}$ channel so 
$\mathrm{pebble}$ is next to this node. The reason of the $\mathrm{since}$ channel is to count how many iterations passed after the node is added to the stack. Therefore, $\mathrm{since}$ channel increases by 1 after the activation.

\begin{align*}
\*X_t & \xleftarrow[]{} \mathrm{concatenate}\left(\*X_t, \*X_{\mathrm{maze}}\right)\\
\forall x, y, \, \*X^{x, y}_{t+1} & \xleftarrow[]{} \sum_{i=0}^2\sum_{j=0}^2 \*W^{i, j}_{\mathrm{DFS}} \odot \*X_t^{x+i-1, y+j-1}\\
\forall i, j, \, \*X_{t+1} & \xleftarrow{} \mathrm{step}\left(\*X^{\mathrm{route}_{i, j}}_{t+1}\right)\\
\*X^{\mathrm{route}}_{t+1} & \leftarrow \mathrm{step}\left(\*X^{\mathrm{route}}_{t+1} + \mathrm{step}\left(\sum_{(i, j)}\*X^{\mathrm{route}_{i, j}}\right)\right)\\[10pt]
\forall \mathrm{chan} \in \{\mathrm{stack},\mathrm{stack\_rank},&\mathrm{stack\_direction}\},\,\\
    \*X^{\mathrm{chan}}_{t+1} &\leftarrow \ReLU\left(\*X^{\mathrm{chan}}\right)\\[10pt]
\mathbf{dbl\_stacked} & \leftarrow \mathrm{sawtooth}_2\left(\*X^{\mathrm{stack}}\right)\\[10pt]
\forall \mathrm{chan} \in \{\mathrm{stack},&\mathrm{stack\_direction}\},\,\\
    \*X^{\mathrm{chan}}_{t+1} &\leftarrow \*X^{\mathrm{chan}}_{t+1} - \*X^{\mathrm{chan}}_t \odot \mathbf{dbl\_stacked}\\[10pt]
\*X^{\mathrm{stack}}_{t+1} &\leftarrow \*X^{\mathrm{stack}}_{t+1} - \left(\*X^{\mathrm{stack}}_t + 1\right) \odot \mathbf{dbl\_stacked}\\
\*X^{\mathrm{pebble}}_{t+1} & \leftarrow \*X^{\mathrm{route}}_{t+1} - \*X^{\mathrm{route}}_t\\
\mathrm{is\_stuck} & \leftarrow  \mathrm{step}\left(\sum_{x, y}\*X^{\mathrm{pebble}, x, y}_{t+1}\right)\\
\mathbf{total\_rank} & \leftarrow \*X^{\mathrm{stack\_rank}}_{t+1} + \*X^{\mathrm{stack\_direction}}_{t+1}\\
\mathbf{total\_rank} & \leftarrow \mathrm{total\_rank} + \mathrm{sawtooth}_0\left(\mathbf{total\_rank}\right) \cdot L\\
\mathrm{min\_total\_rank} &\leftarrow \min_{x, y}\left( \mathbf{total\_rank}^{x, y}_{t+1}\right)\\
\mathbf{is\_popped} &\leftarrow \mathrm{sawtooth}_0\left(\mathbf{total\_rank} - \mathrm{min\_total\_rank}\right)\cdot \mathrm{is\_stuck}\\[10pt]
\forall \mathrm{chan} \in \{\mathrm{stack},\mathrm{stack\_rank},&\mathrm{stack\_direction}\},\,\\
    \*X^{\mathrm{chan}}_{t+1} &\leftarrow \*X^{\mathrm{chan}}_{t+1} - \*X^{\mathrm{chan}}_{t+1} \odot \mathbf{is\_popped}\\
\end{align*}

\subsection{Experimental details}
\label{s:exp_details}

For each experiment (each row in a table), 5 trials were conducted, involving $50,000$ model update steps. 
Each trial took up to $~5$ hours, running on nodes with a single GPU on a High Performance Computing cluster, with GPUs comparable to the RTX 2080 Ti.
In tables, we report standard deviations of each experiment in each metric over all trials. 

\subsection{Extended results}
\label{s:extended results}

In Table~\ref{tab:app_gnn_archs}, we compare Graph Convolutional Networks (GCNs) and Graph Attention Networks (GATs) on the pathfinding task. 
Surprisingly, GATs are outperformed by GCNs on this task, even though the anisotropic capacity of GATs more closely ressembles that of NCAs.
In Table~\ref{tab:app_gnn_grid_rep}, we find that representing the grid-maze as a graph in which only traversable tiles and edges are included in the graph leads to significantly increased performance in the GCN model.



In Table~\ref{tab:app_pf_models} we conduct a hyperparameter sweep over various model architectures.

In Tables~\ref{tab:app_diameter_nca} and \ref{tab:app_pathfinding_nca}, we conduct hyperparameter sweeps on NCA models on the pathfinding and diameter problems, respectively (all prior tables focusing on NCAs comprise select rows from these larger sweeps).


\begin{table}[]
\resizebox{\linewidth}{!}{
\begin{tabular}{|c|r|r|r|r|r|r|r|r|}
\hline
    &       &    &     &      model &                   train & \multicolumn{3}{c|}{test} \\
\cline{5-9}
\cline{5-9}
    &       &    &     &        --- &                   16x16 & \multicolumn{2}{c|}{16x16} &                   32x32 \\
\cline{5-9}
\cline{5-9}
    &       &    &     &  n. params &                accuracy &                accuracy &           pct. complete &                accuracy \\
\textbf{model} & \textbf{\begin{tabular}[r]{@{}r@{}}traversable\\ edges only\end{tabular}} & \textbf{n. layers} & \textbf{n. hid chan} &            &                         &                         &                         &                         \\
\hline
\multirow{18}{*}{\textbf{GCN}} & \multirow{9}{*}{\textbf{False}} & \multirow{3}{*}{\textbf{16}} & \textbf{96 } &     81,600 &            27.22 ± 9.68 &            27.09 ± 9.62 &            10.67 ± 4.10 &             0.04 ± 2.80 \\
    &       &    & \textbf{128} &    143,616 &           15.49 ± 16.35 &           15.41 ± 16.26 &             6.21 ± 6.59 &             0.36 ± 1.28 \\
    &       &    & \textbf{256} &    565,760 &           20.94 ± 17.78 &           20.36 ± 17.29 &             8.85 ± 7.66 &            -1.14 ± 2.69 \\
\cline{3-9}
    &       & \multirow{3}{*}{\textbf{32}} & \textbf{96 } &    158,400 &           27.31 ± 14.44 &           27.13 ± 14.34 &            11.74 ± 6.44 &            -1.52 ± 3.52 \\
    &       &    & \textbf{128} &    278,784 &           28.81 ± 15.18 &           28.54 ± 15.04 &            13.14 ± 7.37 &            -1.43 ± 3.19 \\
    &       &    & \textbf{256} &  1,098,240 &           28.12 ± 19.30 &           27.24 ± 18.71 &           14.92 ± 10.61 &            -4.64 ± 6.94 \\
\cline{3-9}
    &       & \multirow{3}{*}{\textbf{64}} & \textbf{96 } &    312,000 &           17.87 ± 18.31 &           17.87 ± 18.30 &             8.36 ± 8.93 &            -2.02 ± 5.27 \\
    &       &    & \textbf{128} &    549,120 &            7.73 ± 15.80 &            7.69 ± 15.72 &             3.92 ± 8.26 &            -1.65 ± 5.13 \\
    &       &    & \textbf{256} &  2,163,200 &           12.11 ± 19.33 &           11.96 ± 19.09 &            6.66 ± 10.76 &            -0.51 ± 2.43 \\
\cline{2-9}
\cline{3-9}
    & \multirow{9}{*}{\textbf{True }} & \multirow{3}{*}{\textbf{16}} & \textbf{96 } &     81,600 &           50.05 ± 26.80 &           50.16 ± 26.87 &           22.87 ± 13.01 &         -45.97 ± 215.37 \\
    &       &    & \textbf{128} &    143,616 &           56.78 ± 30.19 &           56.97 ± 30.30 &           31.74 ± 17.11 &         -42.62 ± 150.45 \\
    &       &    & \textbf{256} &    565,760 &           55.57 ± 38.38 &           55.59 ± 38.39 &           32.49 ± 22.45 &         -85.07 ± 229.68 \\
\cline{3-9}
    &       & \multirow{3}{*}{\textbf{32}} & \textbf{96 } &    158,400 &           39.26 ± 34.24 &           39.27 ± 34.25 &           19.10 ± 17.12 &           27.99 ± 25.41 \\
    &       &    & \textbf{128} &    278,784 &           45.65 ± 40.03 &           45.68 ± 40.05 &           25.50 ± 22.98 &           33.75 ± 29.60 \\
    &       &    & \textbf{256} &  1,098,240 &  \textbf{69.17} ± 36.46 &  \textbf{68.95} ± 36.35 &  \textbf{40.32} ± 21.27 &  \textbf{50.21} ± 28.55 \\
\cline{3-9}
    &       & \multirow{3}{*}{\textbf{64}} & \textbf{96 } &    312,000 &           39.04 ± 82.35 &           38.42 ± 84.73 &           28.22 ± 17.24 &           16.01 ± 99.26 \\
    &       &    & \textbf{128} &    549,120 &           42.11 ± 37.59 &           42.15 ± 37.61 &           23.21 ± 20.91 &           29.47 ± 29.44 \\
    &       &    & \textbf{256} &  2,163,200 &           51.27 ± 44.19 &           51.26 ± 44.18 &           29.87 ± 25.75 &           44.59 ± 38.42 \\
\hline
\end{tabular}
}
\caption{\textbf{Shortest path problem: graph representation.} Representing the grid-maze as a graph containing only traversable nodes and edges increases generalization.}
\label{tab:app_gnn_grid_rep}
\end{table}

\begin{table}[]
\resizebox{\linewidth}{!}{
\begin{tabular}{|c|r|r|r|r|r|r|r|r|}
\hline
    &       &    &     &      model &                  train & \multicolumn{3}{c|}{test} \\
\cline{5-9}
\cline{5-9}
    &       &    &     &        --- &                  16x16 & \multicolumn{2}{c|}{16x16} &                   32x32 \\
\cline{5-9}
\cline{5-9}
    &       &    &     &  n. params &               accuracy &               accuracy &           pct. complete &                accuracy \\
\textbf{model} & \textbf{positional edge features} & \textbf{n. layers} & \textbf{n. hid chan} &            &                        &                        &                         &                         \\
\hline
\multirow{18}{*}{\textbf{GAT}} & \multirow{9}{*}{\textbf{False}} & \multirow{3}{*}{\textbf{16}} & \textbf{96 } &    153,600 &          40.98 ± 23.59 &          41.20 ± 23.72 &             5.84 ± 5.69 &           31.06 ± 17.73 \\
    &       &    & \textbf{128} &    270,336 &           55.75 ± 4.41 &           56.11 ± 4.56 &            13.11 ± 5.10 &            41.76 ± 3.02 \\
    &       &    & \textbf{256} &  1,064,960 &          55.22 ± 30.90 &          55.38 ± 30.99 &           16.04 ± 12.82 &           41.43 ± 23.17 \\
\cline{3-9}
    &       & \multirow{3}{*}{\textbf{32}} & \textbf{96 } &    307,200 &          24.93 ± 34.14 &          24.99 ± 34.22 &             5.97 ± 9.22 &           19.04 ± 26.08 \\
    &       &    & \textbf{128} &    540,672 &          56.63 ± 31.71 &          56.62 ± 31.70 &           17.92 ± 11.88 &           44.11 ± 24.82 \\
    &       &    & \textbf{256} &  2,129,920 &          62.84 ± 35.22 &          62.78 ± 35.19 &           22.37 ± 13.54 &           51.25 ± 28.74 \\
\cline{3-9}
    &       & \multirow{3}{*}{\textbf{64}} & \textbf{96 } &    614,400 &          53.83 ± 30.27 &          53.85 ± 30.27 &           16.03 ± 11.84 &           42.17 ± 23.61 \\
    &       &    & \textbf{128} &  1,081,344 &          61.71 ± 34.60 &          61.78 ± 34.64 &           21.99 ± 12.56 &           50.76 ± 28.65 \\
    &       &    & \textbf{256} &  4,259,840 &          44.59 ± 41.08 &          44.66 ± 41.15 &           14.23 ± 15.33 &           37.58 ± 34.97 \\
\cline{2-9}
\cline{3-9}
    & \multirow{9}{*}{\textbf{True }} & \multirow{3}{*}{\textbf{16}} & \textbf{96 } &    156,672 &          27.56 ± 25.32 &          27.69 ± 25.45 &             5.24 ± 6.67 &           21.47 ± 19.66 \\
    &       &    & \textbf{128} &    274,432 &          44.97 ± 25.86 &          45.14 ± 25.96 &            12.28 ± 9.92 &           33.99 ± 19.51 \\
    &       &    & \textbf{256} &  1,073,152 &          26.54 ± 36.38 &          26.57 ± 36.41 &             4.17 ± 5.72 &           20.10 ± 27.57 \\
\cline{3-9}
    &       & \multirow{3}{*}{\textbf{32}} & \textbf{96 } &    313,344 &          39.26 ± 36.02 &          39.31 ± 36.06 &           10.10 ± 11.75 &           29.72 ± 27.29 \\
    &       &    & \textbf{128} &    548,864 &          54.86 ± 30.75 &          54.97 ± 30.82 &           19.12 ± 11.04 &           41.53 ± 23.28 \\
    &       &    & \textbf{256} &  2,146,304 &          15.51 ± 34.68 &          15.45 ± 34.55 &            5.58 ± 12.47 &           12.64 ± 28.27 \\
\cline{3-9}
    &       & \multirow{3}{*}{\textbf{64}} & \textbf{96 } &    626,688 &  \textbf{73.62} ± 2.12 &  \textbf{73.77} ± 2.13 &            26.01 ± 5.62 &            58.15 ± 2.16 \\
    &       &    & \textbf{128} &  1,097,728 &           71.93 ± 6.94 &           72.10 ± 6.97 &            26.18 ± 3.82 &           49.45 ± 23.78 \\
    &       &    & \textbf{256} &  4,292,608 &          15.48 ± 34.61 &          15.46 ± 34.58 &             2.50 ± 5.58 &           12.96 ± 28.98 \\
\cline{1-9}
\cline{2-9}
\cline{3-9}
\multirow{9}{*}{\textbf{GCN}} & \multirow{9}{*}{\textbf{False}} & \multirow{3}{*}{\textbf{16}} & \textbf{96 } &    153,600 &          53.07 ± 30.04 &          53.26 ± 30.14 &           26.36 ± 15.27 &           39.91 ± 22.68 \\
    &       &    & \textbf{128} &    270,336 &           68.96 ± 4.56 &           69.19 ± 4.65 &            37.93 ± 4.27 &            51.16 ± 3.53 \\
    &       &    & \textbf{256} &  1,064,960 &          31.27 ± 42.82 &          31.29 ± 42.85 &           18.66 ± 25.55 &           24.07 ± 32.97 \\
\cline{3-9}
    &       & \multirow{3}{*}{\textbf{32}} & \textbf{96 } &    307,200 &          40.64 ± 37.91 &          40.58 ± 37.86 &           19.96 ± 19.50 &           30.95 ± 29.40 \\
    &       &    & \textbf{128} &    540,672 &          44.89 ± 41.90 &          44.95 ± 41.95 &           24.19 ± 23.36 &           34.92 ± 32.78 \\
    &       &    & \textbf{256} &  2,129,920 &          68.75 ± 38.43 &          68.51 ± 38.30 &           39.55 ± 22.11 &           56.53 ± 31.62 \\
\cline{3-9}
    &       & \multirow{3}{*}{\textbf{64}} & \textbf{96 } &    614,400 &          65.21 ± 36.49 &          65.33 ± 36.55 &           35.88 ± 20.19 &           52.38 ± 29.40 \\
    &       &    & \textbf{128} &  1,081,344 &          49.12 ± 44.84 &          49.13 ± 44.86 &           27.60 ± 25.22 &           39.17 ± 35.80 \\
    &       &    & \textbf{256} &  4,259,840 &          69.48 ± 38.85 &          69.35 ± 38.77 &  \textbf{40.27} ± 22.52 &  \textbf{58.19} ± 32.54 \\
\hline
\end{tabular}
}
\caption{\textbf{Shortest path problem -- GNN model architectures} (without weight-sharing): Despite being more closely aligned to the NCA, the GAT does not outperform GCNs, potentially due to insufficient training time.}
\label{tab:app_gnn_archs}
\end{table}

\begin{table}
\resizebox{\linewidth}{!}{

\begin{tabular}{|c|r|r|r|r|r|r|r|}
\hline
    &    &     &       model &                  train & \multicolumn{3}{c|}{test} \\
\cline{4-8}
\cline{4-8}
    &    &     &         --- &                  16x16 & \multicolumn{2}{c|}{16x16} &                  32x32 \\
\cline{4-8}
\cline{4-8}
    &    &     &   n. params &             accuracies &             accuracies &          pct. complete &             accuracies \\
\textbf{model} & \textbf{n. layers} & \textbf{n. hid chan} &             &                        &                        &                        &                        \\
\hline
\multirow{9}{*}{\textbf{GCN}} & \multirow{3}{*}{\textbf{16}} & \textbf{96 } &       9,600 &          47.02 ± 26.29 &          47.09 ± 26.32 &          19.38 ± 10.84 &       -131.51 ± 291.67 \\
    &    & \textbf{128} &      16,896 &          44.59 ± 40.73 &          44.77 ± 40.89 &          25.56 ± 23.36 &       -136.19 ± 169.80 \\
    &    & \textbf{256} &      66,560 &           79.85 ± 1.99 &           79.90 ± 1.97 &           46.31 ± 1.69 &       -194.49 ± 296.77 \\
\cline{2-8}
    & \multirow{3}{*}{\textbf{32}} & \textbf{96 } &       9,600 &          37.89 ± 34.59 &          37.95 ± 34.65 &          18.24 ± 16.65 &          25.01 ± 23.75 \\
    &    & \textbf{128} &      16,896 &          46.41 ± 42.99 &          46.41 ± 42.99 &          26.80 ± 25.24 &          32.57 ± 29.88 \\
    &    & \textbf{256} &      66,560 &          69.61 ± 38.92 &          69.36 ± 38.78 &          41.09 ± 22.97 &          43.92 ± 27.12 \\
\cline{2-8}
    & \multirow{3}{*}{\textbf{64}} & \textbf{96 } &       9,600 &         12.87 ± 110.48 &         11.50 ± 114.14 &          20.54 ± 10.68 &        -20.56 ± 134.55 \\
    &    & \textbf{128} &      16,896 &          35.10 ± 32.34 &          35.19 ± 32.41 &          18.83 ± 17.30 &          19.77 ± 20.87 \\
    &    & \textbf{256} &      66,560 &          33.88 ± 46.40 &          33.98 ± 46.54 &          18.94 ± 26.10 &          29.56 ± 40.55 \\
\cline{1-8}
\cline{2-8}
\multirow{9}{*}{\textbf{MLP}} & \multirow{3}{*}{\textbf{16}} & \textbf{96 } &  16,257,024 &           77.94 ± 1.32 &           14.75 ± 0.68 &            2.73 ± 0.18 &            0.00 ± 0.00 \\
    &    & \textbf{128} &  21,565,440 &           75.90 ± 3.30 &           12.42 ± 1.74 &            2.77 ± 0.10 &            0.00 ± 0.00 \\
    &    & \textbf{256} &  42,799,104 &          52.11 ± 29.83 &            8.13 ± 4.57 &            1.88 ± 1.11 &            0.00 ± 0.00 \\
\cline{2-8}
    & \multirow{3}{*}{\textbf{32}} & \textbf{96 } &  16,257,024 &          31.90 ± 43.68 &            5.85 ± 8.04 &            1.24 ± 1.71 &            0.00 ± 0.00 \\
    &    & \textbf{128} &  21,565,440 &          29.47 ± 40.36 &            4.83 ± 6.63 &            1.02 ± 1.39 &            0.00 ± 0.00 \\
    &    & \textbf{256} &  42,799,104 &           2.82 ± 42.89 &          -8.34 ± 25.59 &            0.54 ± 1.20 &            0.00 ± 0.00 \\
\cline{2-8}
    & \multirow{3}{*}{\textbf{64}} & \textbf{96 } &  16,257,024 &           70.69 ± 3.29 &           13.91 ± 1.04 &            2.74 ± 0.20 &            0.00 ± 0.00 \\
    &    & \textbf{128} &  21,565,440 &           63.58 ± 3.09 &           12.48 ± 1.74 &            2.45 ± 0.20 &            0.00 ± 0.00 \\
    &    & \textbf{256} &  42,799,104 &           52.92 ± 6.63 &           12.82 ± 1.78 &            1.94 ± 0.17 &            0.00 ± 0.00 \\
\cline{1-8}
\cline{2-8}
\multirow{8}{*}{\textbf{NCA}} & \multirow{3}{*}{\textbf{16}} & \textbf{96 } &      86,400 &           99.78 ± 0.05 &           96.04 ± 0.33 &           93.86 ± 0.59 &          62.71 ± 35.01 \\
    &    & \textbf{128} &     152,064 &  \textbf{99.97} ± 0.01 &           96.67 ± 0.20 &           95.16 ± 0.26 &           87.08 ± 1.26 \\
    &    & \textbf{256} &     599,040 &          79.97 ± 44.71 &          78.03 ± 43.62 &          77.58 ± 43.37 &          70.12 ± 39.22 \\
\cline{2-8}
    & \multirow{3}{*}{\textbf{32}} & \textbf{96 } &      86,400 &           99.67 ± 0.28 &           96.79 ± 0.70 &           96.26 ± 0.47 &           84.75 ± 6.69 \\
    &    & \textbf{128} &     152,064 &           99.92 ± 0.09 &  \textbf{97.78} ± 0.32 &           97.61 ± 0.30 &  \textbf{91.78} ± 1.51 \\
    &    & \textbf{256} &     599,040 &          79.68 ± 44.54 &          78.33 ± 43.79 &          78.27 ± 43.76 &          74.24 ± 41.55 \\
\cline{2-8}
    & \multirow{2}{*}{\textbf{64}} & \textbf{128} &     152,064 &           99.39 ± 0.15 &           97.44 ± 0.13 &  \textbf{97.61} ± 0.20 &           90.06 ± 2.93 \\
    &    & \textbf{256} &     599,040 &          79.14 ± 44.24 &          77.90 ± 43.55 &          78.35 ± 43.80 &          73.92 ± 41.36 \\
\hline
\end{tabular}
}
\caption{\textbf{Shortest path problem: model architecture.} Neural Cellular Automata generalize best, while Graph Convolutional Networks generalize better than Multilayer Perceptrons. }
\label{tab:app_pf_models}
\vspace*{-0.5cm}
\end{table}

\begin{table}
\centering
\resizebox{4.7in}{!}{
\Rotatebox{270}{%
    \begin{tabular}{|c|r|r|r|r|r|r|r|r|r|r|r|}
\hline
    &       &   &       &       &       &     &       model & \multicolumn{2}{c|}{train} & \multicolumn{2}{c|}{test} \\
\cline{8-12}
\cline{8-12}
    &       &   &       &       &       &     &         --- & \multicolumn{2}{c|}{16x16} &                  16x16 &                  32x32 \\
\cline{8-12}
\cline{8-12}
    &       &   &       &       &       &     &   n. params &                sol len &             accuracies &             accuracies &             accuracies \\
\textbf{model} & \textbf{evo. data} & \textbf{kernel} & \textbf{max-pool} & \textbf{shared weights} & \textbf{cut corners} & \textbf{n. hid chan} &             &                        &                        &                        &                        \\
\hline
\multirow{32}{*}{\textbf{NCA}} & \multirow{16}{*}{\textbf{False}} & \multirow{8}{*}{\textbf{3}} & \multirow{4}{*}{\textbf{False}} & \multirow{2}{*}{\textbf{False}} & \textbf{False} & \textbf{128} &   9,584,640 &           24.09 ± 0.00 &           99.82 ± 0.06 &           78.52 ± 0.71 &          16.85 ± 14.55 \\
    &       &   &       &       & \textbf{True } & \textbf{128} &   5,324,800 &           24.09 ± 0.00 &           99.85 ± 0.07 &           73.38 ± 2.33 &           12.77 ± 5.20 \\
\cline{5-12}
    &       &   &       & \multirow{2}{*}{\textbf{True }} & \textbf{False} & \textbf{128} &     149,760 &           24.09 ± 0.00 &           95.93 ± 0.42 &           83.65 ± 0.53 &           33.46 ± 9.10 \\
    &       &   &       &       & \textbf{True } & \textbf{128} &      83,200 &           24.09 ± 0.00 &          78.05 ± 43.63 &          66.89 ± 37.40 &          -9.86 ± 12.18 \\
\cline{4-12}
\cline{5-12}
    &       &   & \multirow{4}{*}{\textbf{True }} & \multirow{2}{*}{\textbf{False}} & \textbf{False} & \textbf{128} &   9,584,640 &           24.09 ± 0.00 &           99.79 ± 0.05 &           79.21 ± 1.90 &          30.88 ± 12.92 \\
    &       &   &       &       & \textbf{True } & \textbf{128} &   5,324,800 &           24.09 ± 0.00 &           99.67 ± 0.15 &           82.57 ± 0.93 &           46.06 ± 4.74 \\
\cline{5-12}
    &       &   &       & \multirow{2}{*}{\textbf{True }} & \textbf{False} & \textbf{128} &     149,760 &           24.09 ± 0.00 &          77.41 ± 43.27 &          67.20 ± 37.57 &          27.47 ± 52.59 \\
    &       &   &       &       & \textbf{True } & \textbf{128} &      83,200 &           24.09 ± 0.00 &           95.72 ± 0.45 &           86.54 ± 1.07 &          56.31 ± 26.80 \\
\cline{3-12}
\cline{4-12}
\cline{5-12}
    &       & \multirow{8}{*}{\textbf{5}} & \multirow{4}{*}{\textbf{False}} & \multirow{2}{*}{\textbf{False}} & \textbf{False} & \textbf{128} &  26,624,000 &           24.09 ± 0.00 &  \textbf{99.93} ± 0.04 &           59.30 ± 4.76 &           -8.37 ± 9.73 \\
    &       &   &       &       & \textbf{True } & \textbf{128} &  13,844,480 &           24.09 ± 0.00 &           99.84 ± 0.03 &           68.04 ± 6.29 &          -1.71 ± 25.78 \\
\cline{5-12}
    &       &   &       & \multirow{2}{*}{\textbf{True }} & \textbf{False} & \textbf{128} &     416,000 &           24.09 ± 0.00 &           95.09 ± 0.94 &           77.63 ± 1.01 &           17.58 ± 2.79 \\
    &       &   &       &       & \textbf{True } & \textbf{128} &     216,320 &           24.09 ± 0.00 &           95.06 ± 1.69 &           80.73 ± 1.64 &           26.02 ± 7.79 \\
\cline{4-12}
\cline{5-12}
    &       &   & \multirow{4}{*}{\textbf{True }} & \multirow{2}{*}{\textbf{False}} & \textbf{False} & \textbf{128} &  26,624,000 &           24.09 ± 0.00 &           99.90 ± 0.06 &           67.60 ± 1.17 &            2.78 ± 8.84 \\
    &       &   &       &       & \textbf{True } & \textbf{128} &  13,844,480 &           24.09 ± 0.00 &           99.79 ± 0.14 &           75.09 ± 2.22 &          23.18 ± 12.67 \\
\cline{5-12}
    &       &   &       & \multirow{2}{*}{\textbf{True }} & \textbf{False} & \textbf{128} &     416,000 &           24.09 ± 0.00 &           96.62 ± 0.96 &           76.89 ± 0.93 &          21.85 ± 14.69 \\
    &       &   &       &       & \textbf{True } & \textbf{128} &     216,320 &           24.09 ± 0.00 &           96.30 ± 0.54 &           81.43 ± 0.45 &           49.74 ± 2.58 \\
\cline{2-12}
\cline{3-12}
\cline{4-12}
\cline{5-12}
    & \multirow{16}{*}{\textbf{True }} & \multirow{8}{*}{\textbf{3}} & \multirow{4}{*}{\textbf{False}} & \multirow{2}{*}{\textbf{False}} & \textbf{False} & \textbf{128} &   9,584,640 &           29.45 ± 0.72 &           86.88 ± 0.87 &           88.37 ± 1.58 &          39.32 ± 17.32 \\
    &       &   &       &       & \textbf{True } & \textbf{128} &   5,324,800 &           29.56 ± 0.21 &           76.45 ± 0.97 &           89.88 ± 0.98 &           46.38 ± 2.70 \\
\cline{5-12}
    &       &   &       & \multirow{2}{*}{\textbf{True }} & \textbf{False} & \textbf{128} &     149,760 &           26.34 ± 1.27 &          69.32 ± 38.76 &          71.53 ± 39.99 &          -6.90 ± 42.96 \\
    &       &   &       &       & \textbf{True } & \textbf{128} &      83,200 &           24.81 ± 0.54 &          69.91 ± 39.08 &          72.31 ± 40.43 &         -62.54 ± 95.29 \\
\cline{4-12}
\cline{5-12}
    &       &   & \multirow{4}{*}{\textbf{True }} & \multirow{2}{*}{\textbf{False}} & \textbf{False} & \textbf{128} &   9,584,640 &           30.58 ± 0.58 &           82.15 ± 2.07 &           89.02 ± 1.56 &          55.48 ± 15.45 \\
    &       &   &       &       & \textbf{True } & \textbf{128} &   5,324,800 &           30.24 ± 0.37 &           83.18 ± 2.97 &           89.41 ± 0.86 &  \textbf{70.35} ± 2.62 \\
\cline{5-12}
    &       &   &       & \multirow{2}{*}{\textbf{True }} & \textbf{False} & \textbf{128} &     149,760 &           27.43 ± 0.15 &           87.76 ± 0.79 &           90.02 ± 0.44 &          30.02 ± 64.80 \\
    &       &   &       &       & \textbf{True } & \textbf{128} &      83,200 &           26.19 ± 0.36 &           86.96 ± 1.41 &           89.54 ± 0.91 &          31.89 ± 44.56 \\
\cline{3-12}
\cline{4-12}
\cline{5-12}
    &       & \multirow{8}{*}{\textbf{5}} & \multirow{4}{*}{\textbf{False}} & \multirow{2}{*}{\textbf{False}} & \textbf{False} & \textbf{128} &  26,624,000 &           30.52 ± 0.41 &           82.21 ± 0.81 &  \textbf{90.50} ± 0.21 &           43.19 ± 4.75 \\
    &       &   &       &       & \textbf{True } & \textbf{128} &  13,844,480 &           29.62 ± 0.52 &           80.37 ± 1.31 &           88.96 ± 1.64 &           47.23 ± 3.04 \\
\cline{5-12}
    &       &   &       & \multirow{2}{*}{\textbf{True }} & \textbf{False} & \textbf{128} &     416,000 &           28.03 ± 0.13 &           86.06 ± 1.22 &           88.95 ± 1.36 &           38.04 ± 5.62 \\
    &       &   &       &       & \textbf{True } & \textbf{128} &     216,320 &           27.61 ± 0.27 &           85.77 ± 1.01 &           89.96 ± 1.71 &          29.37 ± 26.44 \\
\cline{4-12}
\cline{5-12}
    &       &   & \multirow{4}{*}{\textbf{True }} & \multirow{2}{*}{\textbf{False}} & \textbf{False} & \textbf{128} &  26,624,000 &  \textbf{30.68} ± 0.20 &           87.87 ± 1.00 &           88.82 ± 0.36 &          40.32 ± 12.18 \\
    &       &   &       &       & \textbf{True } & \textbf{128} &  13,844,480 &           30.24 ± 0.18 &           84.83 ± 2.20 &           88.24 ± 1.02 &           53.65 ± 7.34 \\
\cline{5-12}
    &       &   &       & \multirow{2}{*}{\textbf{True }} & \textbf{False} & \textbf{128} &     416,000 &           28.03 ± 0.35 &           85.54 ± 1.06 &           89.88 ± 0.55 &          48.12 ± 16.01 \\
    &       &   &       &       & \textbf{True } & \textbf{128} &     216,320 &           27.44 ± 0.29 &           86.44 ± 1.09 &           89.58 ± 0.52 &          24.46 ± 29.35 \\
\hline
\end{tabular}
}
}
\caption{\textbf{Diameter problem}: hyperparameter sweep.}
\label{tab:app_diameter_nca}
\end{table}

\begin{table}
\centering
\resizebox{4.6in}{!}{
\centering
\Rotatebox{270}{%
    \begin{tabular}{|c|r|r|r|r|r|r|r|r|r|r|r|r|}
\hline
    &       &   &       &       &       &     &       model & \multicolumn{2}{c|}{train} & \multicolumn{3}{c|}{test} \\
\cline{8-13}
\cline{8-13}
    &       &   &       &       &       &     &         --- & \multicolumn{2}{c|}{16x16} & \multicolumn{2}{c|}{16x16} &                  32x32 \\
\cline{8-13}
\cline{8-13}
    &       &   &       &       &       &     &   n. params &                sol len &             accuracies &             accuracies &          pct. complete &             accuracies \\
\textbf{model} & \textbf{evo. data} & \textbf{kernel} & \textbf{max-pool} & \textbf{shared weights} & \textbf{cut corners} & \textbf{n. hid chan} &             &                        &                        &                        &                        &                        \\
\hline
\multirow{32}{*}{\textbf{NCA}} & \multirow{16}{*}{\textbf{False}} & \multirow{8}{*}{\textbf{3}} & \multirow{4}{*}{\textbf{False}} & \multirow{2}{*}{\textbf{False}} & \textbf{False} & \textbf{128} &   9,732,096 &            9.02 ± 0.00 &           99.92 ± 0.08 &           97.12 ± 0.42 &           97.07 ± 0.59 &           86.72 ± 3.29 \\
    &       &   &       &       & \textbf{True } & \textbf{128} &   5,406,720 &            9.02 ± 0.00 &           99.89 ± 0.11 &           97.01 ± 0.38 &           96.75 ± 0.76 &           86.85 ± 1.66 \\
\cline{5-13}
    &       &   &       & \multirow{2}{*}{\textbf{True }} & \textbf{False} & \textbf{128} &     152,064 &            9.02 ± 0.00 &           98.91 ± 0.69 &           97.38 ± 0.65 &           97.66 ± 0.93 &           90.89 ± 2.09 \\
    &       &   &       &       & \textbf{True } & \textbf{128} &      84,480 &            9.02 ± 0.00 &          79.12 ± 44.24 &          78.09 ± 43.66 &          78.30 ± 43.77 &          73.87 ± 41.30 \\
\cline{4-13}
\cline{5-13}
    &       &   & \multirow{4}{*}{\textbf{True }} & \multirow{2}{*}{\textbf{False}} & \textbf{False} & \textbf{128} &   9,732,096 &            9.02 ± 0.00 &           99.94 ± 0.02 &           96.47 ± 0.13 &           96.07 ± 0.06 &           86.12 ± 2.09 \\
    &       &   &       &       & \textbf{True } & \textbf{128} &   5,406,720 &            9.02 ± 0.00 &           99.91 ± 0.04 &           97.11 ± 0.42 &           96.76 ± 0.57 &           86.30 ± 1.38 \\
\cline{5-13}
    &       &   &       & \multirow{2}{*}{\textbf{True }} & \textbf{False} & \textbf{128} &     152,064 &            9.02 ± 0.00 &           98.96 ± 0.46 &           97.09 ± 0.56 &           97.76 ± 0.41 &           87.00 ± 3.14 \\
    &       &   &       &       & \textbf{True } & \textbf{128} &      84,480 &            9.02 ± 0.00 &           99.58 ± 0.12 &           98.08 ± 0.28 &           98.32 ± 0.31 &           88.61 ± 4.07 \\
\cline{3-13}
\cline{4-13}
\cline{5-13}
    &       & \multirow{8}{*}{\textbf{5}} & \multirow{4}{*}{\textbf{False}} & \multirow{2}{*}{\textbf{False}} & \textbf{False} & \textbf{128} &  27,033,600 &            9.02 ± 0.00 &          79.95 ± 44.69 &          74.95 ± 41.90 &          74.97 ± 41.92 &          51.93 ± 35.75 \\
    &       &   &       &       & \textbf{True } & \textbf{128} &  14,057,472 &            9.02 ± 0.00 &  \textbf{99.95} ± 0.03 &           94.89 ± 0.79 &           95.05 ± 0.61 &           78.46 ± 3.80 \\
\cline{5-13}
    &       &   &       & \multirow{2}{*}{\textbf{True }} & \textbf{False} & \textbf{128} &     422,400 &            9.02 ± 0.00 &          39.70 ± 54.36 &          38.89 ± 53.26 &          39.01 ± 53.41 &          36.43 ± 49.89 \\
    &       &   &       &       & \textbf{True } & \textbf{128} &     219,648 &            9.02 ± 0.00 &          59.48 ± 54.30 &          58.29 ± 53.21 &          58.48 ± 53.39 &          54.40 ± 49.66 \\
\cline{4-13}
\cline{5-13}
    &       &   & \multirow{4}{*}{\textbf{True }} & \multirow{2}{*}{\textbf{False}} & \textbf{False} & \textbf{128} &  27,033,600 &            9.02 ± 0.00 &           99.95 ± 0.04 &           92.95 ± 0.55 &           92.10 ± 1.36 &          64.62 ± 16.19 \\
    &       &   &       &       & \textbf{True } & \textbf{128} &  14,057,472 &            9.02 ± 0.00 &           99.87 ± 0.07 &           93.27 ± 1.19 &           93.09 ± 0.99 &           75.81 ± 3.94 \\
\cline{5-13}
    &       &   &       & \multirow{2}{*}{\textbf{True }} & \textbf{False} & \textbf{128} &     422,400 &            9.02 ± 0.00 &          39.40 ± 53.95 &          38.24 ± 52.36 &          38.53 ± 52.77 &          32.80 ± 44.95 \\
    &       &   &       &       & \textbf{True } & \textbf{128} &     219,648 &            9.02 ± 0.00 &           99.40 ± 0.14 &           97.36 ± 0.37 &           97.67 ± 0.18 &           87.28 ± 4.18 \\
\cline{2-13}
\cline{3-13}
\cline{4-13}
\cline{5-13}
    & \multirow{16}{*}{\textbf{True }} & \multirow{8}{*}{\textbf{3}} & \multirow{4}{*}{\textbf{False}} & \multirow{2}{*}{\textbf{False}} & \textbf{False} & \textbf{128} &   9,732,096 &           28.08 ± 0.21 &           98.73 ± 0.11 &           99.25 ± 0.27 &           98.61 ± 0.96 &           97.91 ± 0.50 \\
    &       &   &       &       & \textbf{True } & \textbf{128} &   5,406,720 &           24.04 ± 8.40 &          78.71 ± 44.00 &          79.61 ± 44.51 &          79.59 ± 44.49 &          78.55 ± 43.91 \\
\cline{5-13}
    &       &   &       & \multirow{2}{*}{\textbf{True }} & \textbf{False} & \textbf{128} &     152,064 &           23.75 ± 1.36 &           96.81 ± 1.54 &           99.29 ± 0.53 &           99.53 ± 0.32 &           98.47 ± 1.01 \\
    &       &   &       &       & \textbf{True } & \textbf{128} &      84,480 &           24.50 ± 1.17 &           97.54 ± 0.88 &  \textbf{99.64} ± 0.12 &  \textbf{99.75} ± 0.07 &  \textbf{98.95} ± 0.40 \\
\cline{4-13}
\cline{5-13}
    &       &   & \multirow{4}{*}{\textbf{True }} & \multirow{2}{*}{\textbf{False}} & \textbf{False} & \textbf{128} &   9,732,096 &           28.06 ± 0.23 &           99.49 ± 0.31 &           99.30 ± 0.35 &           98.77 ± 0.97 &           95.22 ± 3.48 \\
    &       &   &       &       & \textbf{True } & \textbf{128} &   5,406,720 &           26.43 ± 1.21 &           98.77 ± 0.39 &           99.48 ± 0.09 &           99.25 ± 0.34 &           94.82 ± 1.32 \\
\cline{5-13}
    &       &   &       & \multirow{2}{*}{\textbf{True }} & \textbf{False} & \textbf{128} &     152,064 &           19.88 ± 6.13 &          76.83 ± 42.96 &          79.39 ± 44.38 &          79.60 ± 44.50 &          78.43 ± 43.85 \\
    &       &   &       &       & \textbf{True } & \textbf{128} &      84,480 &           22.39 ± 7.48 &          78.07 ± 43.64 &          79.64 ± 44.52 &          79.62 ± 44.51 &          78.95 ± 44.13 \\
\cline{3-13}
\cline{4-13}
\cline{5-13}
    &       & \multirow{8}{*}{\textbf{5}} & \multirow{4}{*}{\textbf{False}} & \multirow{2}{*}{\textbf{False}} & \textbf{False} & \textbf{128} &  27,033,600 &           24.55 ± 8.68 &          79.59 ± 44.49 &          79.44 ± 44.41 &          79.12 ± 44.24 &          76.94 ± 43.02 \\
    &       &   &       &       & \textbf{True } & \textbf{128} &  14,057,472 &  \textbf{28.62} ± 0.19 &           98.79 ± 1.33 &           98.72 ± 0.82 &           98.04 ± 1.73 &           96.19 ± 1.12 \\
\cline{5-13}
    &       &   &       & \multirow{2}{*}{\textbf{True }} & \textbf{False} & \textbf{128} &     422,400 &           25.52 ± 0.30 &           95.90 ± 1.40 &           99.04 ± 0.43 &           99.23 ± 0.54 &           97.80 ± 0.88 \\
    &       &   &       &       & \textbf{True } & \textbf{128} &     219,648 &           22.69 ± 3.80 &           95.29 ± 2.23 &           98.40 ± 1.01 &           98.66 ± 1.20 &           96.13 ± 3.56 \\
\cline{4-13}
\cline{5-13}
    &       &   & \multirow{4}{*}{\textbf{True }} & \multirow{2}{*}{\textbf{False}} & \textbf{False} & \textbf{128} &  27,033,600 &           24.61 ± 0.55 &           99.57 ± 0.09 &           98.67 ± 0.36 &           97.78 ± 0.85 &           88.58 ± 9.19 \\
    &       &   &       &       & \textbf{True } & \textbf{128} &  14,057,472 &           24.21 ± 0.65 &           99.42 ± 0.22 &           99.13 ± 0.05 &           98.53 ± 0.23 &           94.94 ± 1.11 \\
\cline{5-13}
    &       &   &       & \multirow{2}{*}{\textbf{True }} & \textbf{False} & \textbf{128} &     422,400 &           22.07 ± 1.38 &           92.99 ± 4.99 &           95.28 ± 6.31 &           93.76 ± 9.44 &          66.13 ± 64.35 \\
    &       &   &       &       & \textbf{True } & \textbf{128} &     219,648 &           18.43 ± 5.20 &           91.74 ± 4.23 &           97.78 ± 1.14 &           98.40 ± 0.71 &           94.08 ± 3.15 \\
\hline
\end{tabular}
}
}
\caption{\textbf{Pathfinding problem}: hyperparameter sweep.}
\label{tab:app_pathfinding_nca}
\end{table}

\end{document}